\documentclass[a4paper,fleqn]{cas-dc}

\usepackage[numbers]{natbib}

\usepackage{amssymb}
\usepackage{bm}
\usepackage{multirow}
\usepackage{graphicx}
\usepackage{subcaption}
\usepackage{amsmath}
\usepackage{pifont}

\def\tsc#1{\csdef{#1}{\textsc{\lowercase{#1}}\xspace}}
\tsc{WGM}
\tsc{QE}

\begin{document}
\ExplSyntaxOn
\cs_gset:Npn \__first_footerline: {}
\ExplSyntaxOff

\makeatletter
\renewcommand{\printorcid}{}
\makeatother

\let\WriteBookmarks\relax
\def\floatpagepagefraction{1}
\def\textpagefraction{.001}

\shorttitle{fMRI2Face}    

\shortauthors{Huo et~al.}

\title [mode = title]{
fMRI2Face: A Full-HD fMRI-Video Dataset and Geometry-Guided Neural Decoding Framework for Dynamic Human Face Reconstruction
}  

\author[1]{Jingyang Huo}
\ead{jyhuo22@m.fudan.edu.cn}

\author[2]{Xiangru Huang}%
\ead{huangxiangru@westlake.edu.cn}

\author[3]{Chentao Shen}

\author[4]{Yikai Wang}

\author[1]{Yun Wang}

\author[1]{Jianxiong Gao}

\author[5]{Shihao Jin}

\author[1]{Yanwei Fu}
\ead{yanweifu@fudan.edu.cn}
\cormark[1]

\author[1]{Jianfeng Feng}

\affiliation[1]{organization={Fudan University},
            addressline={220 Handan Road, Yangpu District},
            city={Shanghai},
            postcode={200433}, 
            country={China}}
\affiliation[2]{organization={Westlake University},
            addressline={No. 18 Shilongshan Road, Xihu District},
            city={Hangzhou},
            postcode={310024}, 
            state={Zhejiang},
            country={China}}
\affiliation[3]{organization={Zhejiang University},
            addressline={No. 866 Yuhangtang Road, Xihu District},
            city={Hangzhou},
            postcode={310058},
            state={Zhejiang},
            country={China}}
\affiliation[4]{organization={Nanyang Technological University},
            addressline={50 Nanyang Avenue},
            city={Singapore},
            postcode={639798},
            country={Singapore}}

\affiliation[5]{organization={Xmov},
            city={Shanghai},
            country={China}}

\cortext[1]{Corresponding author}

\begin{abstract}
Reconstructing dynamic human faces from brain activity provides a powerful way to study how the mind perceives identity, expression, and facial motion.
However, progress in fMRI-based face decoding has been limited by scarce controlled, high-resolution neural datasets and by methods that struggle to recover both identity-specific appearance and time-varying facial dynamics.
We present \textbf{fMRI-Face}, the first fMRI dataset paired with controllable full-HD digital human facial videos rendered at 1920$\times$1080 resolution.
During scanning, participants watched photorealistic, background-free facial videos with controlled identity, expression, and head pose, while fMRI activity was recorded. The resulting dataset contains 62,856 paired fMRI--video samples, providing a structured resource for studying dynamic face perception and reconstruction.
Building on this dataset, we propose \textbf{fMRI2Face}, a geometry-guided neural video decoding framework for reconstructing facial videos from fMRI signals.
fMRI2Face derives two complementary neural controls from brain activity: \textbf{Brain-derived Appearance Context}, which captures global identity-related visual attributes, and \textbf{Morphable 3D Facial Control}, which provides explicit geometry-aware guidance for pose, expression, and non-rigid facial dynamics.
These controls are integrated through \textbf{Neural-Controlled Video Diffusion} with auxiliary latent completion, enabling high-fidelity facial video reconstruction directly from brain activity.
Together, fMRI-Face and fMRI2Face establish a controlled platform for studying dynamic face perception and provide a new benchmark for fMRI-based digital human reconstruction.
\end{abstract}

\begin{keywords}
Neural Decoding \sep Digital Human \sep Face Reconstruction \sep Video Generation
\end{keywords}

\maketitle

\section{Introduction}\label{sec:intro}
How can we glimpse the face someone sees in their mind without ever seeing it ourselves?
From paint to prose to photography, humanity has long sought to externalize internal perception.
Today, neural decoding takes this pursuit further: reconstructing perceived visual experiences directly from brain activity. By mapping neural signals to sensory representations, this emerging field provides a computational bridge between the brain and the visual world, offering a potential “digital window” into human perception.

Recent advances in deep generative modeling have enabled the reconstruction of static natural images and scenes from fMRI signals with remarkable fidelity~\cite{chen2023seeing, ozcelik2023natural, scotti2024mindeye2, huo2024neuropictor}. 
Yet, decoding human \textit{faces}, the most socially and perceptually salient stimuli, remains one of the most challenging frontiers. 
Faces combine highly structured 3D geometry, identity-specific details, and rapid, non-rigid expressive motion. 
Capturing all these aspects from fMRI requires far more than reproducing visual appearance. 
It demands recovering facial dynamics and individuality, the very properties that make each face perceptually distinct.

Progress toward this goal has been limited by two intertwined obstacles.
\textbf{(1) Data scarcity:} existing fMRI-face datasets are typically low-resolution, rely on synthetic or constrained stimuli~\cite{vanrullen2019reconstructing}, or entangle faces with complex, distracting backgrounds~\cite{dado2022hyperrealistic, chen2024fmri}, making it difficult to isolate neural responses for identity, expression, and pose.
\textbf{(2) Methodological gaps:} most prior approaches treat the task as a direct fMRI-to-pixel mapping or use GAN inversion~\cite{vanrullen2019reconstructing, dado2022hyperrealistic, ren2024brain} without explicitly modeling facial structure, often producing blurry, inconsistent, and semantically unstable reconstructions.
Without explicit facial geometry priors or motion-aware facial controls, these models struggle to preserve identity across time or recover faithful facial dynamics, leaving the true “face in the mind” still out of reach.

\begin{figure}
  \centering
  \includegraphics[width=0.85\linewidth]{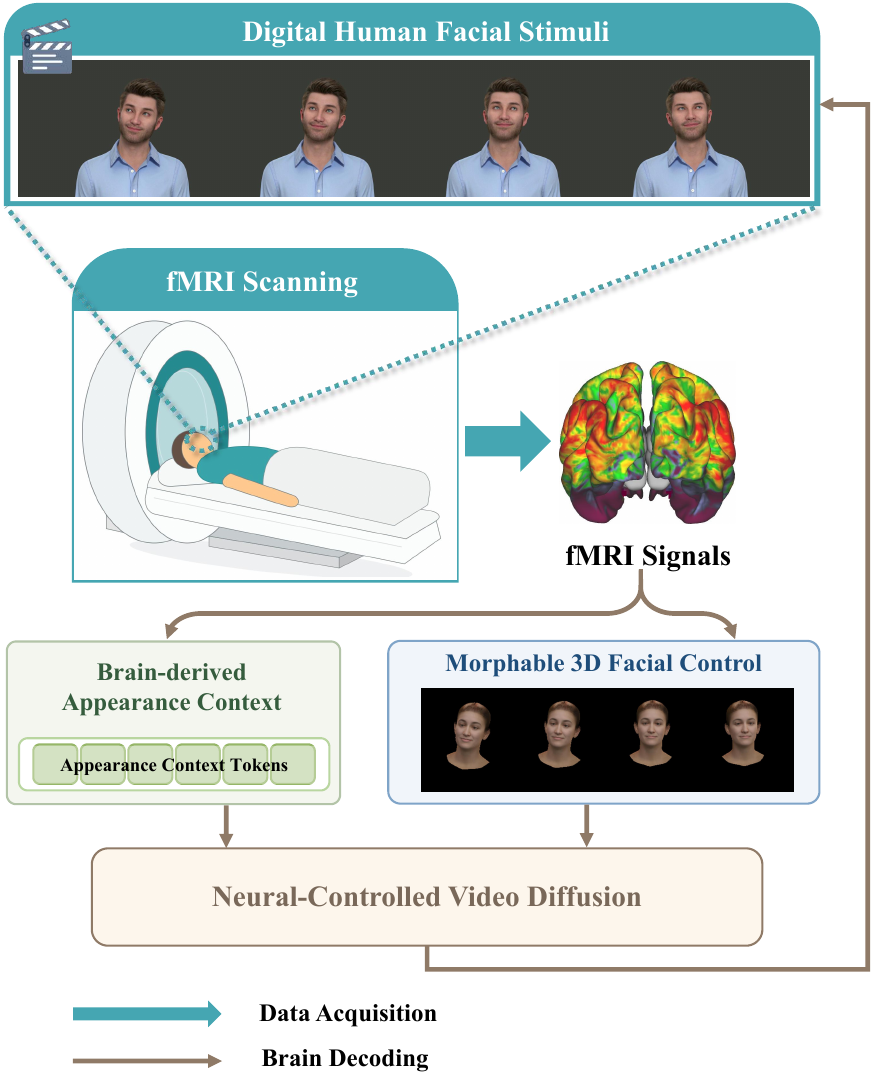}
  \caption{Illustration of the fMRI-Face dataset and the fMRI2Face decoding framework.
  \textbf{Top:} Participants view digital human facial stimuli during fMRI scanning, producing brain responses.
  \textbf{Bottom:} Given recorded fMRI signals, fMRI2Face derives two complementary neural controls: Brain-derived Appearance Context for global visual appearance and Morphable 3D Facial Control for geometry-aware structure and motion.
  The two controls are integrated through Neural-Controlled Video Diffusion to reconstruct facial videos from brain activity.
}
  \label{fig:teaser}
\end{figure}

To overcome the data limitation, we introduce \textbf{fMRI-Face}, a large-scale dataset designed to reveal how the brain encodes dynamic, photorealistic human faces. It is the first fMRI dataset paired with \emph{controllable digital human stimuli} rendered in full HD (1920$\times$1080). 
Fig.~\ref{fig:teaser} provides a schematic view of this data acquisition process, where full-HD digital human facial stimuli are presented during fMRI scanning and paired with the recorded brain responses.
During scanning, participants viewed thousands of background-free facial clips featuring diverse identities, natural expressions, and head movements.
Importantly, unlike prior datasets~\cite{vanrullen2019reconstructing,dado2022hyperrealistic} that use low-resolution/low-quality or cluttered imagery, each stimulus in fMRI-Face is a background-free, parameterized 3D facial sequence rendered under consistent lighting and physically accurate shading. 
This design reduces nuisance variations from background, illumination, and camera artifacts, making perceptual variations primarily reflect facial identity, expression, and pose.
In total, fMRI-Face contains 62,856 paired samples of facial video and fMRI responses, exceeding existing face-decoding datasets in both scale and controllability (Tab.~\ref{tab:dataset_comparison}). This high-fidelity resource provides a powerful foundation for modeling the neural basis of dynamic face perception and reconstructing faces directly from brain activity.

Building on this dataset, we propose \textbf{fMRI2Face}, a geometry-guided neural face video decoding framework that reconstructs photorealistic and temporally coherent facial videos from fMRI signals.
As illustrated in Fig.~\ref{fig:teaser}, our key idea is to integrate global appearance guidance inferred from brain activity with explicit geometry-aware facial dynamics control, thereby connecting neural responses with controllable facial video reconstruction.

Specifically, fMRI2Face learns \textbf{Brain-derived Appearance Context}, a compact set of fMRI-conditioned context tokens that provide global visual cues, including identity-related facial attributes, skin tone, hairstyle, and overall appearance.
These tokens replace the text condition of a pretrained video diffusion model, enabling appearance guidance to be inferred directly from brain activity and maintained consistently across frames.

However, global appearance context alone is not sufficient for faithful dynamic face reconstruction. Expression transitions, head movements, and local non-rigid deformations require structured spatial and temporal guidance. 
Relying only on context tokens therefore tends to produce plausible facial appearance but weakly constrained motion dynamics.
To address this challenge, we introduce \textbf{Morphable 3D Facial Control}, which maps fMRI signals into a structured parametric facial representation. The predicted parameters describe facial geometry, expression, pose, illumination, and image-space transformations, and are rendered through a pretrained morphable face model~\cite{deca} into geometry-consistent facial controls. Rather than serving as photorealistic outputs themselves, these renderings provide explicit spatial and motion guidance over time.

Finally, the two fMRI-derived controls are integrated within \textbf{Neural-Controlled Video Diffusion} for controllable facial video synthesis.
The Brain-derived Appearance Context guides global identity-related appearance and visual style, while the Morphable 3D Facial Control anchors the generation process to fMRI-decoded facial structure, expression, and head motion.
To preserve complete latent-space context beyond the rendered facial regions, we further introduce auxiliary latent completion during video diffusion.
By combining these components, the complete fMRI2Face pipeline reconstructs facial videos that better preserve identity, expression continuity, and temporally coherent facial dynamics directly from brain activity.

\begin{table}
\caption{Comparison of datasets used for neural decoding in fMRI studies, focusing on faces and natural scene videos. }
\label{tab:dataset_comparison}
\centering
\setlength{\tabcolsep}{3pt}
\resizebox{1.0\linewidth}{!}{
\begin{tabular}{lcccc}
\toprule
\textbf{Dataset} & 
\textbf{Stimuli} & 
\textbf{Category} & 
\textbf{Resolution} & 
\textbf{Train\&Test Pairs}
 \\
\midrule
VanRullen2019 \cite{vanrullen2019reconstructing} & Image & Face & 128$\times$128 & 32000 \\
HYPER~\cite{dado2022hyperrealistic} & Image & Face & 224$\times$224 & 3000 \\
\midrule
Wen2018 \cite{wen2018neural} & Video & Nat. Scene & 800$\times$600 & 20000 \\
fMRI-FCVID \cite{lichong_video} & Video & Nat. Scene & 256$\times$256 & 1800 \\
fMRI-WebVid \cite{lichong_video} & Video & Nat. Scene & 596$\times$336 & 7000 \\
NFED \cite{chen2024fmri} & Video & Face & 768$\times$768 & 18000 \\
\midrule
 \textbf{fMRI-Face} & Video & Face & 1920$\times$1080 & 62856 \\
\bottomrule
\end{tabular}
}
\end{table}

\noindent \textbf{Contributions.} We summarize our contributions:
\begin{itemize}
    \item \textbf{fMRI-Face Dataset:} We introduce the first large-scale fMRI dataset paired with controllable, full-HD (1920$\times$1080) digital human facial videos. Its background-free and parameterized 3D facial stimuli provide precise control over identity, expression, and pose, resulting in 62,856 fMRI--video pairs for studying dynamic face perception and reconstruction.
    
    \item \textbf{Neural Face Video Decoding Framework:} We propose \textbf{fMRI2Face}, a neural decoding framework that reconstructs photorealistic and temporally coherent facial videos from fMRI signals by jointly modeling global facial appearance and explicit facial dynamics.
    
    \item \textbf{Brain-derived Appearance Context:} We learn compact fMRI-conditioned context tokens that provide global visual appearance guidance for video generation, enabling the model to preserve identity-related and individual facial attributes over time.
    
    \item \textbf{Morphable 3D Facial Control:} We introduce a structured 3D facial control stream that maps fMRI signals into parametric facial representations. The rendered controls provide geometry-consistent spatial and motion guidance, improving expression continuity and facial motion fidelity.
    
    \item \textbf{Benchmark and Insights:} We establish a new benchmark for fMRI-based dynamic face reconstruction and demonstrate the benefits of combining 3D parametric modeling with diffusion-based video synthesis.
\end{itemize}

\section{Related Work}
\label{sec:related}
 
\subsection{Neural Decoding}
Neural decoding aims to interpret or reconstruct perceptual experiences from brain signals, typically using fMRI for its ability to non-invasively capture whole-brain responses. Progress in this field has been driven by the availability of high-quality datasets. Early studies collected fMRI responses to simple stimuli such as binary patterns~\cite{miyawaki2008visual} and grayscale characters~\cite{schoenmakers2013linear, van2010neural}, enabling foundational work on brain-signal decoding. Later, more complex datasets emerged to better reflect natural perception, including Generic Object Decoding~\cite{horikawa2017generic}, THINGS~\cite{THINGSdata}, BOLD5000~\cite{chang2019bold5000}, and the Natural Scenes Dataset (NSD)~\cite{allen2022massive}, the largest and most detailed fMRI dataset with over 73,000 image presentations.
Alongside data advances, decoding methods progressed from traditional linear models to deep generative approaches~\cite{shen2019deep, mozafari2020reconstructing, ren2021reconstructing, gu2022neurogen}. Most recently, diffusion models~\cite{rombach2022high} have driven rapid improvement in fMRI-to-image reconstruction~\cite{chen2023seeing, zeng2023controllable, ferrante2023brain, fang2024alleviating, qian2023semantic, qian2023fmri}, with recent works~\cite{ozcelik2023natural, scotti2023reconstructing, huo2024neuropictor, scotti2024mindeye2} achieving strong semantic and spatial fidelity.
Beyond static images, several studies have begun decoding dynamic visual stimuli from fMRI~\cite{wen2018neural, lichong_video, chen2023cinematic, gong2024neuroclips}, showing early promise. However, due to the complexity of facial perception and limitations in dataset scale and control, face reconstruction remains notably less consistent than natural-scene decoding.

\subsection{fMRI-to-face Reconstruction}
Face reconstruction from fMRI is more fine-grained than natural scene decoding. The most widely used dataset for this task is collected by VanRullen et al.~\cite{vanrullen2019reconstructing}, which includes fMRI responses from four participants viewing about 8,000 synthetic face images. These images were generated using a VAE-GAN at 128$\times$128 resolution. Several studies~\cite{david2021localizing, chang2022facial, ren2024brain} have built upon this dataset, typically adopting a GAN-inversion framework and aligning fMRI signals with the GAN latent code. However, due to the low resolution and limited realism of the original stimuli, the reconstructed faces often appear blurry, lacking semantic coherence and identity fidelity.
Subsequent datasets such as HYPER~\cite{dado2022hyperrealistic}, which used faces generated by PGGAN~\cite{karras2017progressive}, and NFED~\cite{chen2024fmri}, which includes fMRI-face video pairs, offer higher-resolution stimuli but remain limited in the number of samples. Moreover, these datasets often entangle facial information with background and contextual visual features, making it difficult to isolate and reconstruct facial identity.
In contrast, our fMRI-Face dataset provides high-resolution, background-free digital humans with precise control over identity, expression, and head pose. This design allows us to decode facial appearance and motion free from environmental interference, leading to more accurate, identity-consistent, and perceptually faithful reconstructions.

\subsection{Video Diffusion Models}
Recent video diffusion models have significantly improved the realism and temporal consistency of generated videos~\cite{ho2022video, chen2024videocrafter2, guo2023animatediff}. Early approaches primarily extended image diffusion models to video generation through image-to-video synthesis and temporal U-Net architectures~\cite{ho2022video}, while more recent methods increasingly adopt diffusion transformers (DiT) for scalable spatiotemporal modeling and controllable video generation~\cite{yang2025cogvideox, wan2025wan}. Beyond text-conditioned generation, many works further introduce additional spatial or motion controls, such as poses, depth maps, masks, trajectories, or reference images, to stabilize temporal dynamics and guide video generation~\cite{wang2024motionctrl, vace}. These studies suggest that global conditioning alone is often insufficient for precise motion generation, especially for temporally coherent facial dynamics~\cite{xu2025hunyuanportrait}. Inspired by this observation, our work combines fMRI-conditioned context tokens with geometry-aware facial controls, enabling identity-preserving and motion-consistent face video reconstruction directly from brain activity.

\subsection{Facial Modeling}
Accurate face modeling is essential for graphics and vision tasks. 2D methods synthesize facial animations from images~\cite{fadm,fomm,siarohin2021motion,vid2vid}, but they often lack interpretability and consistency under changes in pose or expression. 
In contrast, 3D parametric face models~\cite{flame,qian2024gaussianavatars,HFavatar,chu2024gpavatar,avatar3r,deng2024portrait4d,deng2024portrait4dv2} provide compact, controllable representations of shape, expression, and pose, enabling stable rendering across motions. 
In this work, we estimate 3D facial parameters directly from fMRI signals and combine morphable 3D decoding with diffusion-based appearance synthesis to achieve identity-preserving recovery of facial motion and appearance from brain activity.

\begin{figure*}
  \begin{centering}
\includegraphics[width=0.99\linewidth]{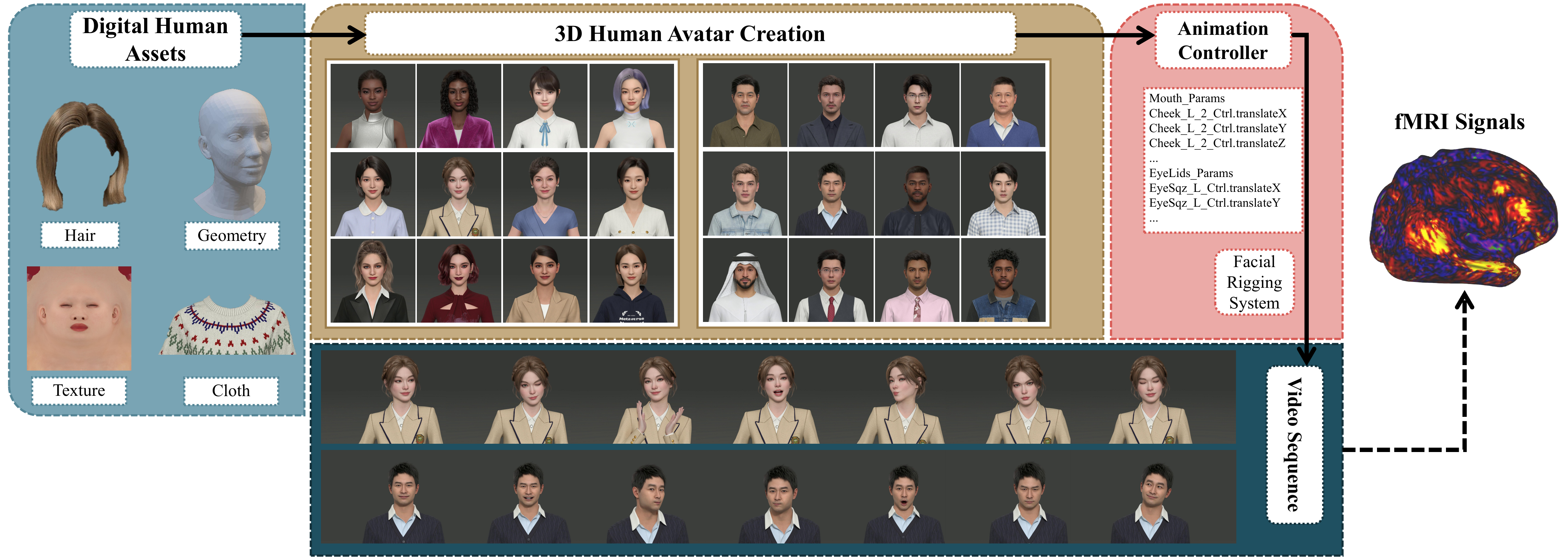}
\par\end{centering}
  \caption{
  Overview of the fMRI-Face dataset workflow. Starting from 3D digital-human creation using modular digital assets, a facial rigging system is configured for each character. Predefined animation controllers drive facial expressions over time, and rendering software produces photorealistic videos for each digital human. Participants view these videos during fMRI scanning.
  }
  \label{fig:stim_paradigm}
\end{figure*}

\section{fMRI-Face Dataset}
\label{sec:data}

In this section, we introduce fMRI-Face, a new dataset of fMRI-digital human facial stimuli. An overview of the full fMRI-Face workflow is provided in Fig.~\ref{fig:stim_paradigm}. We first describe the pipeline for generating 3D digital humans and their corresponding video stimuli (Sec.~\ref{sec:digital_human}). Then, we detail the fMRI acquisition process, including the experimental design, MRI scanning protocol, and preprocessing steps (Sec.~\ref{sec:fmri_scan}). Finally, we use fMRI-Face to analyze the relationship between brain regions and facial stimuli (Sec.~\ref{sec:data_ana}).

\subsection{Digital Human Video Generation Pipeline}
\label{sec:digital_human}
The digital human video generation pipeline in fMRI-Face is supported by photorealistic and large-scale 3D digital human assets, including face geometry, physically-based rendering (PBR) skin textures, styled hair strands, eyebrow, eyelash, cloth, animation sequences and facial rigging system. In this work, we first create thousands of high-quality and animatable 3D digital human avatars by composing 3D assets, such as face geometry, hairstyles and cloth geometry. We then apply a number of facial animation sequences to drive the facial rigging system in rendering software such as Maya and Unreal Engine 5, resulting in one digital human video for each human avatar. The exact group of assets used to create each human avatar is recorded.

\subsubsection{3D Digital Human Creation}
To create a 3D Digital Human, we first decide the face geometry, represented by a mesh $\mathcal{M}=(V, F)$. The faces $F$ are fixed and copied from an artist-created template face mesh, similar to 3DMM~\cite{blanz2023morphable}. The vertex positions $V \in \mathbb{R}^{N \times 3}$ are determined by sampling from a gender-neutral face Principal Component Analysis (PCA) model derived from a large set of scanned face geometry. 

With the face geometry determined, we  fit it to a body mesh. To simplify the experimental setting, we use a male fixed body mesh and a female fixed body mesh for all videos in fMRI-Face. We then apply the PBR skin texture to give the face mesh a photorealistic appearance. Then, we randomly choose hair styles, cloth geometry, eyebrow, eyelash, and run a series of deterministic retargeting algorithms to fit eyes, teeth and all the assets to the face geometry. The retargeting algorithms make minimal but necessary adjustments in the assets to ensure all assets fit to the face geometry correctly, without generating unrealistic artifacts. 

Finally, we use a facial rigging system generation algorithm to generate a suitable facial rigging system for the human avatar. The facial rigging system generation algorithm is an algorithm that maps a face geometry $\mathcal{M}=(V, F)$ to a set of $L$ blend shapes $B=\{ V_j \in \mathbb{R}^{N \times 3} | j = 1, \ldots, L \}$. The face animation can be represented by linear blending, i.e., given a vector of blend weights $\boldsymbol{w} \in [0, 1]^L$, the deformed face can be written as
\begin{equation}
    V(\boldsymbol{w}) = V + \sum_{j=1}^L w_j (V_j - V). \label{eq:rigging}
\end{equation}
The facial rigging system generation algorithm is a data-driven algorithm powered by a dataset of scanned facial rigging systems $\{(V_i, B_i)\}$, which intuitively interpolates scanned rigging system data to generate a realistic rigging system for the human avatar. 

\subsubsection{Animation and Video Generation}
Given the generated digital human, we apply an animation sequence $A=\{\boldsymbol{a}_t | t=1, \ldots, T\}$ to the facial rigging system. At time step $t$, a predefined controller function takes the animation signal $\boldsymbol{a}_t$ and outputs a blend weight vector $\boldsymbol{w}_t = \textup{Controller}(\boldsymbol{a}_t)$, which generates the face animation according to equation $\eqref{eq:rigging}$. To construct the animation dataset, we extract animation segments from the template animation sequences and bind them to different digital human identities. In total, we generated 2,174 digital human videos, each lasting 8 seconds, including 1,193 female and 981 male avatars. We further annotate each video with multiple attributes, including clothing category, hairstyle category, a corresponding canonical portrait rendered with neutral expression, frontal pose, and uniform lighting, as well as motion configuration. 
More dataset details are provided in the supplementary material.

\subsection{fMRI Dataset and Preprocessing}
\label{sec:fmri_scan}

We collected fMRI data from three subjects (ages 22–24; two females and one male). The experimental procedures were approved by the ethical review board, and informed consent was obtained from all participants.

\subsubsection{Experimental Design}
Visual stimuli were constructed using digital human videos generated as described in Sec.~\ref{sec:digital_human}. Each clip consisted of 240 frames at 30 fps. In total, we obtained 2,174 unique video clips, with 180 held out as the test set and the remaining 1,994 used for training. 
During fMRI scanning, each clip was presented as part of a stimulus trial consisting of an 8-second digital human video followed by a 4-second inter-trial interval (ITI). A red central fixation dot was displayed during the ITI to promote stable gaze fixation. Each run comprised 24 such trials, resulting in a total duration of 312 seconds per run.
Each participant completed 10 fMRI sessions, totaling 97 runs per subject. A high-resolution T1-weighted anatomical image was collected during the first session for structural reference and alignment. All participants viewed the full set of 2,174 video clips, presented in randomized order. 
Among these, 154 clips were repeated once to assess neural response reliability, while the remaining clips were presented only once.

\subsubsection{MRI Acquisition}
All scans were performed on a 3T MRI scanner equipped with a 32-channel RF head coil. Participants viewed the LCD screen ($8^\circ \times 8^\circ$) via a mirror mounted on the RF coil, with fixation guided by a red central dot ($0.4^\circ \times 0.4^\circ$). High-resolution structural images were collected using a T1-weighted MPRAGE sequence (0.8-mm isotropic resolution, repetition time (TR) = 2500 ms, echo time (TE) = 2.22 ms, flip angle $8^{\circ}$), where TR denotes the time interval between successive acquisitions and TE denotes the time between excitation and signal readout. Functional images were acquired using a gradient-echo echo-planar imaging (EPI) sequence with whole-brain coverage (2-mm isotropic resolution, TR = 800 ms, TE = 37 ms, flip angle $52^{\circ}$, multi-band factor = 8). Each task-fMRI run consisted of 390 volumes.

\subsubsection{MRI Pre-processing}
The MRI data were preprocessed using the standardized \texttt{fMRIPrep} pipeline~\cite{fmriprep1, fmriprep2} and registered onto the 32k\_fs\_LR surface space. Following established practices in visual decoding~\cite{mind3d, huo2024neuropictor, cinebrain}, we focused on visual regions of interest (ROIs) defined by the Human Connectome Project Multi-Modal Parcellation (HCP-MMP). The ROIs included V1, V2, V3, V3A, V3B, V3CD, V4, LO1, LO2, LO3, PIT, V4t, V6, V6A, V7, V8, PH, FFC, IP0, MT, MST, FST, VVC, VMV1, VMV2, VMV3, PHA1, PHA2, PHA3, TE2p, and IPS1, comprising 8,921 vertices in total. We performed z-score normalization across vertices within each run.

\subsubsection{Data Split}
We use 1,994 videos for training and hold out 180 videos for testing. Because some videos were viewed twice, the final train set contains 2,012 fMRI--digital human facial pairs and the test set contains 316 pairs. Since functional images were acquired with a repetition time (TR) of 0.8 s, each 8-second video segment spans 10 fMRI volumes (TRs). Each 8-second video is divided into 9 overlapping clips using a sliding window with a step size of 1 TR and a duration of 2 TRs (1.6 s). This preprocessing yields 18,108 fMRI--video clips for training and 2,844 for testing per subject.
Aggregating data across all subjects, the fMRI-Face dataset contains 62,856 fMRI--video samples for training and evaluation.

\subsection{Neural Response Reliability}
\label{sec:data_ana}

\begin{figure}
\centering
\begin{subfigure}{1.0\linewidth}
  \includegraphics[width=1.0\linewidth]{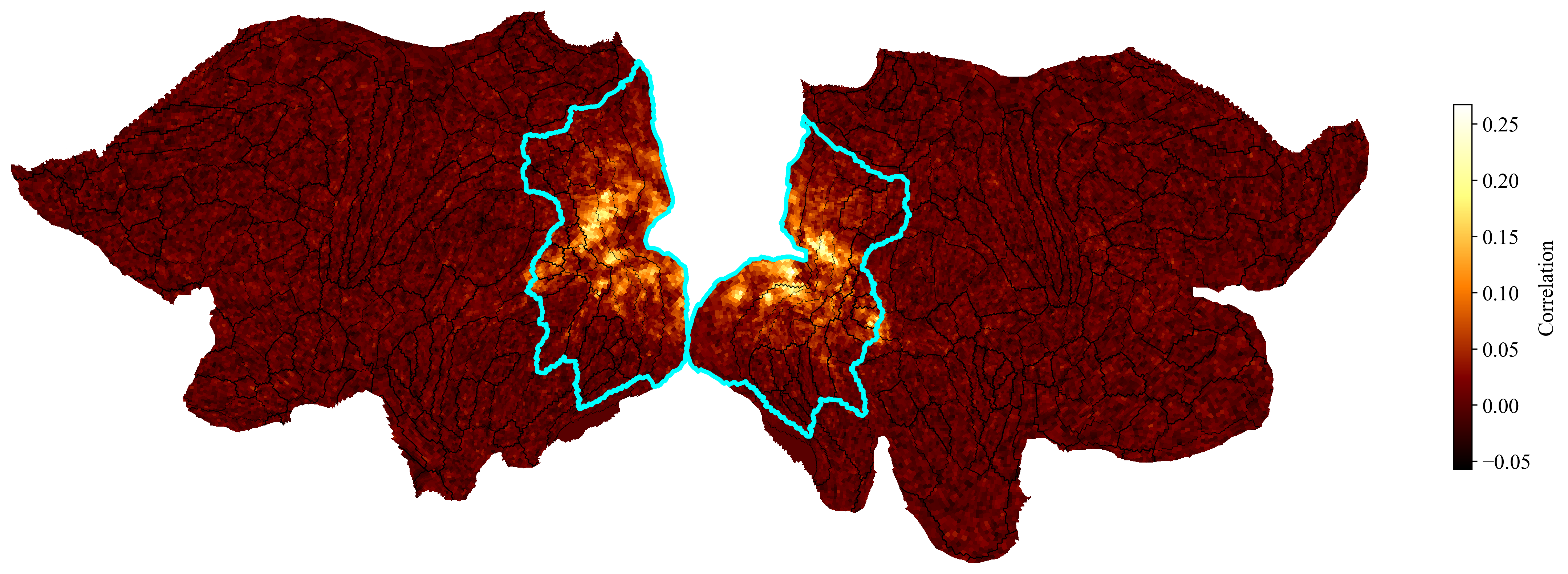}
  \caption{}
  \label{fig:corr_heatmap}
\end{subfigure}
\vfill
\begin{subfigure}{0.48\linewidth}
  \includegraphics[width=1.0\linewidth]{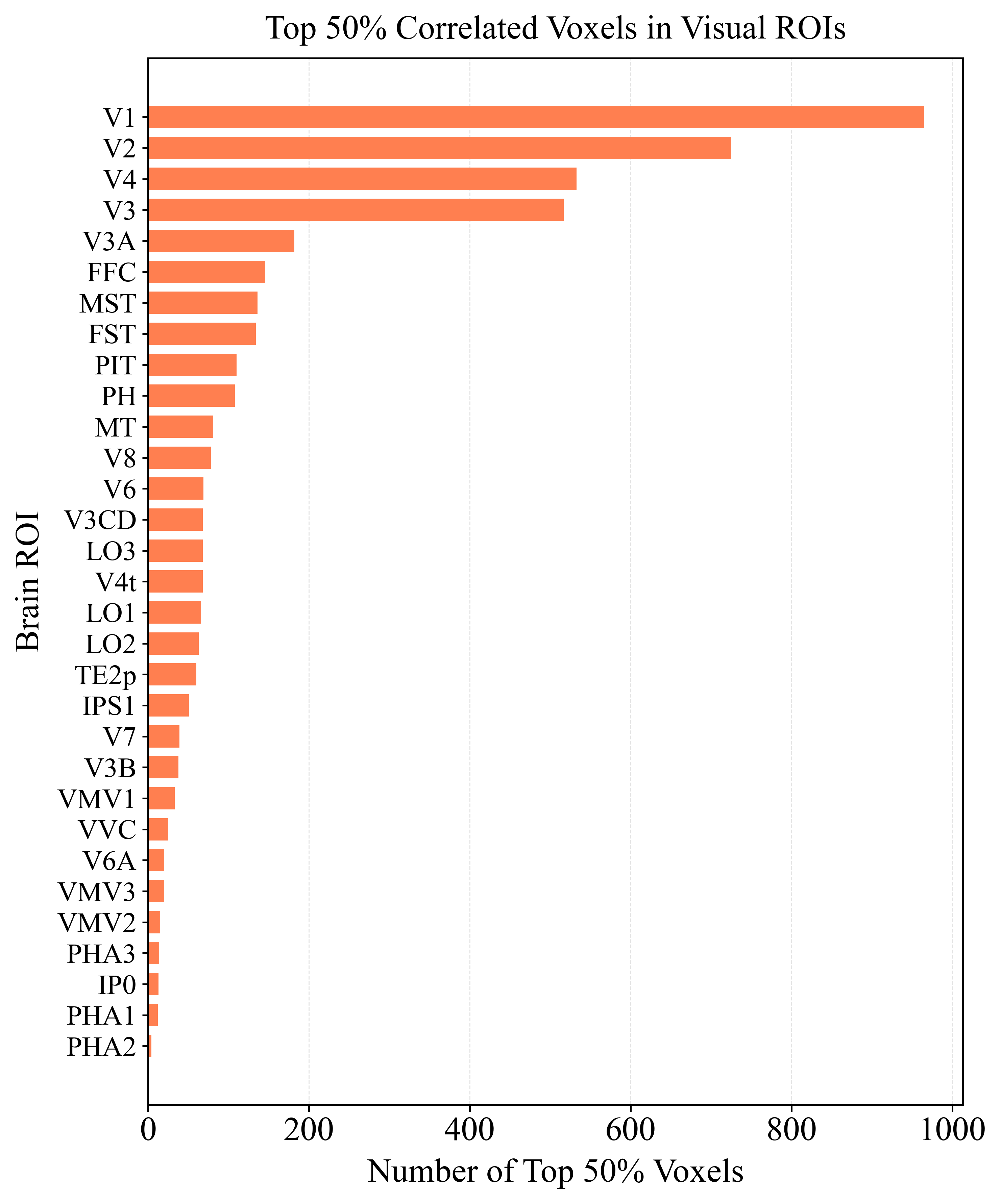}
  \caption{}
  \label{fig:top50_num}
\end{subfigure}
\hfill
\begin{subfigure}{0.48\linewidth}
  \includegraphics[width=1.0\linewidth]{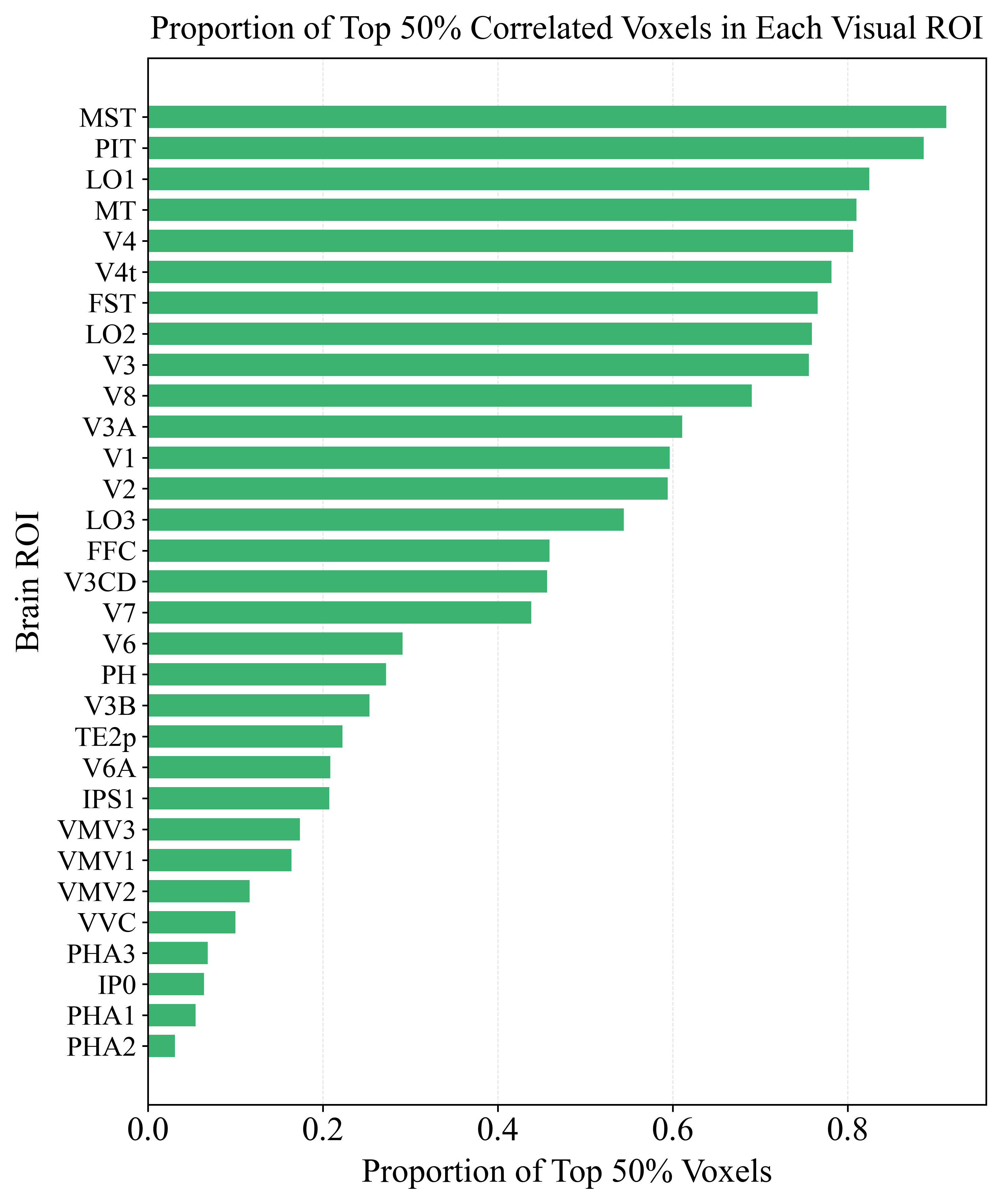}
  \caption{}
  \label{fig:top50_ratio}
\end{subfigure}
\caption{
(a) Neural reliability heatmap across the cortex based on repeated video clips; cyan contours denote selected VC ROIs and black contours denote HCP region boundaries.
(b) Number of top 50\% correlated vertices in each visual ROI. 
(c) Proportion of top 50\% correlated vertices in each visual ROI.
}
\label{fig:combined_results}
\end{figure}

To assess the quality and reliability of fMRI signals, we computed vertex-wise Pearson correlations between fMRI responses from two repeated trials of digital-human video clips. Fig.~\ref{fig:corr_heatmap} shows the average correlation map across subjects. Cyan contours indicate predefined visual ROIs introduced in Sec.~\ref{sec:fmri_scan}. We observe that vertices with highest consistency are primarily within these visual ROIs. 

We analyzed the consistency of brain responses across 31 predefined visual ROIs during repeated viewings of the same video stimuli. Specifically, within the 8,921 vertices comprising the visual ROIs, we selected the top 50\% with the highest vertex-wise Pearson correlations across trials. For each ROI, we report (1) the number of vertices in this top 50\% subset (Fig.~\ref{fig:top50_num}) and (2) the proportion of top-50\% vertices relative to total vertices (Fig.~\ref{fig:top50_ratio}).

Our analysis reveals a distinction between early and higher-level visual areas. The Primary Visual Cortex (V1) and early visual areas (V2, V3, V4) contribute the greatest number of highly consistent vertices (Fig.~\ref{fig:top50_num}), consistent with their high reliability in previous research~\cite{allen2022massive,gong2023large}. However, the highest proportion of consistent vertices within each ROI is found in mid- and high-level areas, such as MT (Middle Temporal), MST (Medial Superior Temporal), LO1 (Lateral Occipital), PIT (Posterior Inferotemporal complex), and FFC (Fusiform Face Complex). These areas are associated with motion perception (MT, MST), object representation (LO, PIT), and face processing (FFC). Notably, the FFC ranks among the top ROIs in vertex-wise consistency, supporting its involvement in perception of the digital human faces used in our stimuli.

\begin{figure*}
  \centering
  \includegraphics[width=0.98\linewidth]{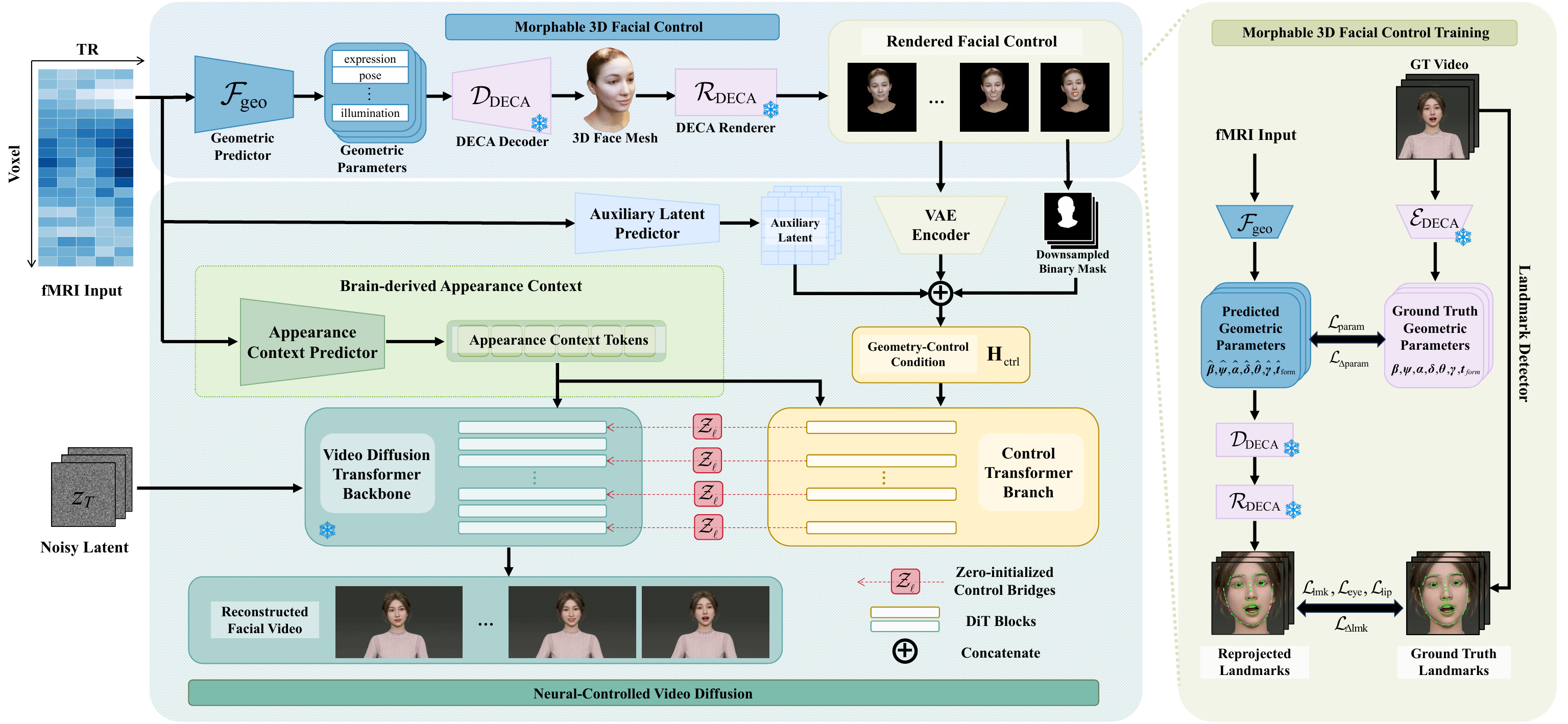}
  \caption{
  Overview of the fMRI2Face framework.
  \textbf{Left:} Given an fMRI input, \textbf{Morphable 3D Facial Control} predicts DECA parameters through the Geometric Predictor $\mathcal{F}_{\mathrm{geo}}$, decodes them into a 3D facial mesh, and renders a geometry-consistent facial guidance sequence $\hat{\mathbf{I}}_{r}$.
  In parallel, \textbf{Brain-derived Appearance Context} predicts context tokens $\mathbf{C}_{a}$ that provide global visual appearance guidance.
  \textbf{Neural-Controlled Video Diffusion} integrates $\mathbf{C}_{a}$ with the geometry-control condition constructed from the rendered-control latent, auxiliary latent, and spatial mask, and injects the control through a parallel transformer branch with zero-initialized residual bridges to reconstruct facial videos.
  \textbf{Right:} The Morphable 3D Facial Control stream is trained using parameter-level and landmark-based reprojection losses derived from ground-truth videos.
  }
  \label{fig:pipeline}
\end{figure*}

\section{fMRI2Face Framework}
\label{sec:method}
In this section, we present \textbf{fMRI2Face}, a neural face video decoding framework that reconstructs a digital human's appearance and motion from brain activity. We formulate this task as translating spatiotemporal fMRI responses into facial video clips, where the generated video should preserve both the global visual appearance of the perceived face and temporal facial dynamics, including expression transitions and head movements.

\subsection{Problem Setup}
Given paired fMRI recordings and digital human facial videos, our goal is to reconstruct the perceived facial video clip from the corresponding neural responses. 
Each target clip spans 1.6 seconds and is represented as a 24-frame video sequence sampled at 15 fps, denoted as $\mathbf{V} = \{ I_i \}_{i=1}^{24}$. 
This clip contains both global facial appearance, such as identity-related visual attributes, and time-varying facial dynamics, such as head motion and non-rigid expression changes.

Since the blood-oxygen-level-dependent (BOLD) response is delayed relative to stimulus onset, the fMRI signal corresponding to a visual event is observed several seconds later. Rather than assuming a single fixed hemodynamic offset, we associate each 2-TR video clip with a delayed fMRI response window spanning 5--9 TRs after the clip onset. Specifically, for a clip, we define the fMRI input as
$\mathbf{F} = [\mathbf{x}_{5}, \mathbf{x}_{6}, \ldots, \mathbf{x}_{9}]
    \in \mathbb{R}^{T \times d},$
where $T=5$ denotes the number of delayed fMRI measurements and $d=8{,}921$ denotes the number of cortical vertices sampled from the selected visual ROIs.

The decoding task is formulated as learning a mapping
$\mathcal{G}_{\Theta}: \mathbb{R}^{T \times d} \rightarrow \mathbb{R}^{K \times H \times W \times 3},$
parameterized by $\Theta$, which maps the fMRI sequence $\mathbf{F}$ to the reconstructed facial video:
\begin{equation}
    \hat{\mathbf{V}} = \mathcal{G}_{\Theta}(\mathbf{F}).
\end{equation}
The reconstructed video $\hat{\mathbf{V}}$ is expected to preserve the global facial appearance and recover temporally coherent facial dynamics that correspond to the perceived stimulus.

\subsection{Overview}

As illustrated in Fig.~\ref{fig:pipeline}, fMRI2Face reconstructs facial videos through two complementary neural conditioning streams: \textbf{Brain-derived Appearance Context} for global visual appearance, and \textbf{Morphable 3D Facial Control} for geometry-aware facial dynamics.
The two streams are integrated through \textbf{Neural-Controlled Video Diffusion} to generate high-fidelity facial videos from brain activity.
Specifically, the framework consists of three parts:

(i) \textbf{Brain-derived Appearance Context.}
We learn compact fMRI-conditioned context tokens that encode global visual appearance cues, including identity-related facial attributes, skin tone, hairstyle, and overall appearance.

(ii) \textbf{Morphable 3D Facial Control.}
We predict parametric facial attributes from fMRI signals and instantiate them as a 3D facial mesh $\mathbf{M}$, capturing the perceived facial geometry, pose, and expression.
The mesh is rendered into a 2D facial guidance image $\mathbf{I}_{r}$, which provides geometry-consistent spatial and motion guidance.

(iii) \textbf{Neural-Controlled Video Diffusion.}
The two fMRI-derived controls are integrated within a pretrained video diffusion model for controllable facial video synthesis.
In this stage, a learnable auxiliary latent representation is introduced to provide additional fMRI-derived context beyond the morphable facial rendering.
Through this unified pipeline, fMRI2Face preserves both global identity and dynamic facial motion in the reconstructed videos.

\subsection{Brain-derived Appearance Context}
\label{sec:app}
The first conditioning stream extracts global visual appearance information from fMRI signals and maps it into the conditioning space of the video diffusion model.
Instead of using a text prompt, we learn Brain-derived Appearance Context, a set of fMRI-conditioned context tokens that encode clip-level appearance cues, including identity-related facial attributes, skin tone, hairstyle, and overall appearance.

\subsubsection{Appearance Context Predictor}
Diffusion-based generators are typically conditioned on text embeddings that guide the generated visual content.
Prior works~\cite{wang2025visual, faceadapter} in natural image generation and face reenactment have shown that replacing text conditions with visual embeddings can effectively preserve appearance and identity.
Inspired by this, we design an appearance context predictor that maps fMRI signals to context tokens compatible with the pretrained video diffusion transformer.

Given an fMRI input window $\mathbf{F} \in \mathbb{R}^{T \times d}$, where $T=5$ and $d=8{,}921$, we first project the fMRI measurements into the model conditioning space and add temporal positional encodings to the projected tokens.
A transformer decoder with 64 learnable query tokens of dimension 4096 then extracts appearance-related information:
\begin{equation}
    \mathbf{C}_{a} = \mathcal{F}_{\mathrm{app}}(\mathbf{F}),
\end{equation}
where $\mathbf{C}_{a} \in \mathbb{R}^{64 \times 4096}$ denotes the resulting Brain-derived Appearance Context, and $\mathcal{F}_{\mathrm{app}}$ denotes the appearance context predictor.
This representation summarizes the fMRI window into appearance conditions for subsequent video generation.
The context tokens are then injected into the pretrained diffusion transformer through its original context-conditioning interface, ensuring compatibility with the pretrained video diffusion backbone.

\subsubsection{Training}
To optimize the Brain-derived Appearance Context, we adopt the latent-space flow-matching objective used in Wan~\cite{wan2025wan}, which aligns the predicted velocity with the target latent flow.
Let $\mathbf{z}_{0}$ denote the latent representation of the target video clip obtained from the frozen VAE.
For each training step, we sample a diffusion time $\tau$ and construct the noisy latent $\mathbf{z}_{\tau}$ and target velocity $\mathbf{v}_{\tau}$ following the scheduler.
The appearance context predictor is optimized by:
\begin{equation}
    \mathcal{L}_{\mathrm{app}} =
    \mathbb{E}_{\tau}
    \left[
    \left\|
    \mathbf{v}_{\theta}(\mathbf{z}_{\tau}, \tau, \mathbf{C}_{a})
    -
    \mathbf{v}_{\tau}
    \right\|_2^2
    \right],
\end{equation}
where $\mathbf{v}_{\theta}$ denotes the model-predicted latent velocity.
During this stage, the VAE and DiT backbone remain frozen, and only $\mathcal{F}_{\mathrm{app}}$ is trained.
This encourages the learned appearance context $\mathbf{C}_{a}$ to align fMRI signals with the latent video generation space of the pretrained diffusion model.

The resulting appearance context provides global guidance for the whole video clip, preserving identity-related appearance and visual style over time, while fine-grained facial motion is handled by the subsequent Morphable 3D Facial Control stream.

\subsection{Morphable 3D Facial Control}
\label{sec:deca}
While the Brain-derived Appearance Context provides global visual guidance, it does not explicitly constrain time-varying facial geometry.
To provide structured motion guidance, we introduce Morphable 3D Facial Control, which maps fMRI signals into parametric facial representations and renders them into geometry-consistent facial controls.

\subsubsection{Parametric Facial Representation} 
We adopt DECA~\cite{deca}, a pretrained morphable face model, as the structured facial representation. Given a facial image $\mathbf{I}$, DECA encodes it into a set of interpretable parameters, including shape $\boldsymbol{\beta}$, expression $\boldsymbol{\psi}$, pose $\boldsymbol{\theta}$, albedo $\boldsymbol{\alpha}$, detail $\boldsymbol{\delta}$, illumination $\boldsymbol{\gamma}$, and image transformation $\boldsymbol{t}_{\mathrm{form}}$:
\begin{equation}
\{ 
\boldsymbol{\beta}, \boldsymbol{\psi}, \boldsymbol{\alpha}, 
\boldsymbol{\delta}, \boldsymbol{\theta}, \boldsymbol{\gamma},
\boldsymbol{t}_{\mathrm{form}}
\}
=
\mathcal{E}_{\mathrm{DECA}}(\mathbf{I}), 
\end{equation}
where $\mathcal{E}_{\mathrm{DECA}}$ denotes the DECA encoder.

The DECA decoder $\mathcal{D}_{\mathrm{DECA}}$ reconstructs a 3D facial mesh $\mathbf{M}$, which can be rendered into the image plane by the differentiable renderer $\mathcal{R}_{\mathrm{DECA}}$:
\begin{equation}
\begin{aligned}
\mathbf{M}
&=
\mathcal{D}_{\mathrm{DECA}}
(
\boldsymbol{\beta}, \boldsymbol{\psi},
\boldsymbol{\alpha}, \boldsymbol{\delta},
\boldsymbol{\theta}, \boldsymbol{\gamma}
), \\
\mathbf{I}_{r}
&=
\mathcal{R}_{\mathrm{DECA}}
(
\mathbf{M},
\boldsymbol{\theta},
\boldsymbol{\gamma},
\boldsymbol{t}_{\mathrm{form}}
).
\end{aligned}
\end{equation}

\subsubsection{Geometric Predictor}
Given an fMRI input sequence $\mathbf{F}$, our goal is to predict a sequence of DECA parameters corresponding to the perceived facial video.
We use a temporal Transformer encoder to model the fMRI sequence and aggregate the temporal information with a class token $\mathbf{z}_{\mathrm{cls}}$.

Specifically, the Geometric Predictor first predicts a first-frame reference state from $\mathbf{z}_{\mathrm{cls}}$, including
$\hat{\boldsymbol{\beta}}_{\mathrm{ref}}$,
$\hat{\boldsymbol{\psi}}_{\mathrm{ref}}$,
$\hat{\boldsymbol{\alpha}}_{\mathrm{ref}}$,
$\hat{\boldsymbol{\delta}}_{\mathrm{ref}}$,
$\hat{\boldsymbol{\theta}}_{\mathrm{ref}}$,
$\hat{\boldsymbol{\gamma}}_{\mathrm{ref}}$, and
$\hat{\boldsymbol{t}}_{\mathrm{form},\mathrm{ref}}$.
This state corresponds to the first frame of the clip and serves as a stable anchor for identity-related shape, appearance-related attributes, and the initial facial configuration.

To model temporal facial dynamics, we further predict frame-wise residuals for temporally varying parameters, including expression ($\hat{\boldsymbol{\psi}}$), pose ($\hat{\boldsymbol{\theta}}$), detail ($\hat{\boldsymbol{\delta}}$), and image transformation ($\hat{\boldsymbol{t}}_{\mathrm{form}}$).
The first-frame parameters are given by the reference state, while subsequent frames are obtained by adding predicted residuals to this reference:
\begin{equation}
\begin{aligned}
\hat{\boldsymbol{\psi}}_{k}
&=
\hat{\boldsymbol{\psi}}_{\mathrm{ref}}
+
\Delta \hat{\boldsymbol{\psi}}_{k}, \\
\hat{\boldsymbol{\theta}}_{k}
&=
\hat{\boldsymbol{\theta}}_{\mathrm{ref}}
+
\Delta \hat{\boldsymbol{\theta}}_{k}, \\
\hat{\boldsymbol{\delta}}_{k}
&=
\hat{\boldsymbol{\delta}}_{\mathrm{ref}}
+
\Delta \hat{\boldsymbol{\delta}}_{k}, \\
\hat{\boldsymbol{t}}_{\mathrm{form},k}
&=
\hat{\boldsymbol{t}}_{\mathrm{form},\mathrm{ref}}
+
\Delta \hat{\boldsymbol{t}}_{\mathrm{form},k},
\end{aligned}
\quad k=2,\ldots,K .
\end{equation}
For parameters expected to remain stable within a clip, including shape, albedo, and illumination, we replicate the reference prediction across frames.
This residual formulation anchors the clip at the first facial state and focuses temporal modeling on expression, pose, and other motion-related variations, while reducing unnecessary fluctuations in stable facial attributes.

The full prediction process is written as:
\begin{equation}
\{
\hat{\boldsymbol{\beta}},
\hat{\boldsymbol{\psi}},
\hat{\boldsymbol{\alpha}},
\hat{\boldsymbol{\delta}},
\hat{\boldsymbol{\theta}},
\hat{\boldsymbol{\gamma}},
\hat{\boldsymbol{t}}_{\mathrm{form}}
\}
=
\mathcal{F}_{\mathrm{geo}}(\mathbf{F}),
\end{equation}
where $\mathcal{F}_{\mathrm{geo}}$ denotes the \textbf{Geometric Predictor}, which maps fMRI signals to a sequence of morphable 3D facial parameters.
We then convert the predicted parameters into rendered facial control images.
For each frame $k$, the predicted parameters are decoded into a 3D facial mesh:
\begin{equation}
\hat{\mathbf{M}}_{k}
=
\mathcal{D}_{\mathrm{DECA}}
(
\hat{\boldsymbol{\beta}}_{k},
\hat{\boldsymbol{\psi}}_{k},
\hat{\boldsymbol{\alpha}}_{k},
\hat{\boldsymbol{\delta}}_{k},
\hat{\boldsymbol{\theta}}_{k},
\hat{\boldsymbol{\gamma}}_{k}
).
\end{equation}
The mesh is then rendered into the image plane as:
\begin{equation}
\hat{\mathbf{I}}_{r,k}
=
\mathcal{R}_{\mathrm{DECA}}
(
\hat{\mathbf{M}}_{k},
\hat{\boldsymbol{\theta}}_{k},
\hat{\boldsymbol{\gamma}}_{k},
\hat{\boldsymbol{t}}_{\mathrm{form},k}
),
\quad k=1,\ldots,K .
\end{equation}
The resulting rendered sequence
$\hat{\mathbf{I}}_{r}=\{\hat{\mathbf{I}}_{r,k}\}_{k=1}^{K}$
serves as geometry-consistent facial guidance for video generation, providing explicit spatial and motion cues over time.

\subsubsection{Training}
We supervise the Morphable 3D Facial Control stream with both parameter-level and reprojection-level losses.
The parameter-level losses constrain the predicted DECA parameters, while the reprojection-level losses encourage the rendered facial geometry to align with the target facial motion.

\noindent
\textit{Parameter-level supervision.}
For each target video, we extract DECA parameters from the ground-truth frames as supervision. This encourages accurate prediction of facial attributes.
The parameter regression loss is defined as:
\begin{equation}
\mathcal{L}_{\mathrm{param}}
=
\sum_{k=1}^{K}
\sum_{x \in \{
\boldsymbol{\beta},
\boldsymbol{\psi},
\boldsymbol{\alpha},
\boldsymbol{\delta},
\boldsymbol{\theta},
\boldsymbol{\gamma},
\boldsymbol{t}_{\mathrm{form}}
\}}
\left\|
\hat{x}_{k} - x_{k}
\right\|_2^2 .
\end{equation}

To further encourage temporally consistent facial dynamics, we introduce a temporal difference loss on expression and pose, which are most directly related to dynamic facial motion. Given a parameter sequence $x=\{x_k\}_{k=1}^{K}$, its first-order temporal difference is:
\begin{equation}
\Delta x_k = x_k - x_{k-1},
\quad k=2,\ldots,K .
\end{equation}
The temporal parameter loss is:
\begin{equation}
\mathcal{L}_{\Delta\mathrm{param}}
=
\sum_{x \in \{\boldsymbol{\psi}, \boldsymbol{\theta}\}}
\sum_{k=2}^{K}
\left\|
\Delta \hat{x}_k - \Delta x_k
\right\|_2^2 .
\end{equation}
This term encourages the predicted motion trajectories to follow the temporal evolution of the perceived facial video.

\noindent
\textit{Reprojection-level supervision.}
We apply landmark-based reprojection losses between the predicted facial geometry and the target face.
Let $\mathbf{p}_{k,i}$ and $\hat{\mathbf{p}}_{k,i}$ denote the target and predicted 2D landmark positions of the $i$-th landmark at frame $k$, respectively.
The landmark reprojection loss is:
\begin{equation}
\mathcal{L}_{\mathrm{lmk}}
=
\sum_{k=1}^{K}
\sum_{i=1}^{N_{\mathrm{lmk}}}
\left\|
\mathbf{p}_{k,i}
-
\hat{\mathbf{p}}_{k,i}
\right\|_1 .
\end{equation}
We also introduce a temporal landmark difference loss to provide explicit supervision on landmark motion over time:
\begin{equation}
\mathcal{L}_{\Delta\mathrm{lmk}}
=
\sum_{k=2}^{K}
\sum_{i=1}^{N_{\mathrm{lmk}}}
\left\|
(\hat{\mathbf{p}}_{k,i}-\hat{\mathbf{p}}_{k-1,i})
-
(\mathbf{p}_{k,i}-\mathbf{p}_{k-1,i})
\right\|_1 .
\end{equation}

To maintain natural mouth articulation, we constrain the relative offsets between upper and lower lip landmarks:
\begin{equation}
\mathcal{L}_{\mathrm{lip}}
=
\sum_{k=1}^{K}
\sum_{(i,j)\in \mathcal{P}_{\mathrm{lip}}}
\left\|
(\hat{\mathbf{p}}_{k,i} - \hat{\mathbf{p}}_{k,j})
-
(\mathbf{p}_{k,i} - \mathbf{p}_{k,j})
\right\|_1 ,
\end{equation}
where $\mathcal{P}_{\mathrm{lip}}$ denotes the set of upper--lower lip landmark pairs.

To preserve realistic eye openness, we similarly constrain relative offsets between eyelid landmarks:
\begin{equation}
\mathcal{L}_{\mathrm{eye}}
=
\sum_{k=1}^{K}
\sum_{(i,j)\in \mathcal{P}_{\mathrm{eye}}}
\left\|
(\hat{\mathbf{p}}_{k,i} - \hat{\mathbf{p}}_{k,j})
-
(\mathbf{p}_{k,i} - \mathbf{p}_{k,j})
\right\|_1 ,
\end{equation}
where $\mathcal{P}_{\mathrm{eye}}$ denotes the set of eyelid landmark pairs.

\noindent
\textit{Final objective.}
The total training loss combines all terms with balancing coefficients:
\begin{equation}
\mathcal{L}_{\mathrm{geo}}
=
\lambda_{\mathrm{param}} \mathcal{L}_{\mathrm{param}}
+
\lambda_{\mathrm{lmk}} \mathcal{L}_{\mathrm{lmk}}
+
\lambda_{\mathrm{lip}} \mathcal{L}_{\mathrm{lip}}
+
\lambda_{\mathrm{eye}} \mathcal{L}_{\mathrm{eye}}
+
\mathcal{L}_{\Delta},
\end{equation}
where
$\mathcal{L}_{\Delta}
=
\lambda_{\Delta\mathrm{param}}
\mathcal{L}_{\Delta\mathrm{param}}
+
\lambda_{\Delta\mathrm{lmk}}
\mathcal{L}_{\Delta\mathrm{lmk}}$.
This joint supervision encourages accurate parameter prediction, stable geometry, and coherent facial motion from fMRI signals.

\subsection{Neural-Controlled Video Diffusion}
\label{sec:dualcond}

The previous two streams provide complementary neural conditions for facial video reconstruction.
The Brain-derived Appearance Context $\mathbf{C}_{a}$ provides global visual guidance, while the Morphable 3D Facial Control provides temporally aligned geometry and motion cues.
To integrate these conditions into video generation, we build \textbf{Neural-Controlled Video Diffusion} on a pretrained video diffusion transformer.
The pretrained denoising transformer serves as the main generation backbone, while an additional geometry-control branch injects fMRI-derived facial controls through residual conditioning.
This design preserves the generative prior of the pretrained video model while providing a dedicated pathway for geometry-aware facial motion control.

\subsubsection{Condition Construction}
The appearance context $\mathbf{C}_{a}$ is injected through the original context-conditioning interface of the pretrained video diffusion transformer, replacing the prompt-derived text context. This allows the denoising backbone to receive global appearance information directly from fMRI signals.

In parallel, the Morphable 3D Facial Control is converted into a latent video control condition.
Let $\hat{\mathbf{I}}_{r}$ denote the rendered facial guidance sequence obtained from the fMRI-predicted morphable 3D parameters.
We encode $\hat{\mathbf{I}}_{r}$ with the frozen video VAE to obtain the rendered-control latent:
\begin{equation}
    \mathbf{H}_{r}
    =
    \mathcal{E}_{\mathrm{VAE}}(\hat{\mathbf{I}}_{r}).
\end{equation}

Since the rendered facial guidance mainly describes the reconstructed face region and does not fully cover visual factors outside the rendered face, we further introduce an auxiliary latent representation:
\begin{equation}
    \mathbf{H}_{\mathrm{aux}}
    =
    \mathcal{F}_{\mathrm{aux}}(\mathbf{F}),
\end{equation}
where $\mathcal{F}_{\mathrm{aux}}$ denotes the auxiliary latent predictor that maps fMRI signals into an auxiliary latent representation.
This auxiliary latent provides additional fMRI-derived context for visual factors not sufficiently represented by the morphable facial rendering, such as non-face regions.

The final control condition is constructed by concatenating the rendered-control latent, the auxiliary latent, and a binary mask sequence:
\begin{equation}
    \mathbf{H}_{\mathrm{ctrl}}
    =
    \mathrm{Concat}
    \left(
    \mathbf{H}_{r},
    \mathbf{H}_{\mathrm{aux}},
    \mathrm{Down}(\mathbf{S}_{r})
    \right),
\end{equation}
where concatenation is performed along the channel dimension.
Here, $\mathbf{S}_{r}$ is a binary mask sequence obtained from $\hat{\mathbf{I}}_{r}$ by setting background pixels to 0 and non-background pixels to 1.
It serves as a spatial indicator of the valid rendered facial-control region at each frame.

\subsubsection{Geometry-Control Injection}

To inject the geometry-control condition into the video generator, we introduce a parallel control branch whose building blocks follow the same transformer architecture as the pretrained denoising backbone.
The control condition $\mathbf{H}_{\mathrm{ctrl}}$ is first mapped into the hidden space of the diffusion transformer by a linear projection, producing control features aligned with the noisy video latents at the same spatiotemporal resolution.
These features are then processed by a subset of transformer blocks copied from the pretrained backbone to progressively refine geometry-aware control representations before residual injection.

At each selected layer $\ell \in \mathcal{J}_{\mathrm{ctrl}}$, the control branch produces a control feature $\mathbf{g}_{\ell}$, which is projected and added to the hidden state of the main denoising branch:
\begin{equation}
    \mathbf{h}_{\ell}
    \leftarrow
    \mathbf{h}_{\ell}
    +
    \mathcal{Z}_{\ell}
    \left(
    \mathbf{g}_{\ell}
    \right),
    \quad \ell \in \mathcal{J}_{\mathrm{ctrl}} .
\end{equation}
where $\mathbf{g}_{\ell}$ denotes the control feature produced by the copied transformer block $\mathcal{B}_{\ell}$ in the geometry-control branch, $\mathbf{h}_{\ell}$ denotes the hidden state of the main denoising branch, and $\mathcal{Z}_{\ell}$ is the residual projection that maps the control feature back to the main branch.

\subsubsection{Training}

We initialize the geometry-control branch from a VACE checkpoint~\cite{vace} and train it with LoRA adaptation.
During this stage, the pretrained denoising backbone, VAE, and appearance context predictor $\mathcal{F}_{\mathrm{app}}$ are frozen.
We optimize the LoRA parameters of the geometry-control branch and the auxiliary latent predictor $\mathcal{F}_{\mathrm{aux}}$.

Given a target facial video $\mathbf{V}$, we encode it into a latent representation $\mathbf{z}_{0}$ using the frozen video VAE.
We sample a diffusion time $\tau$ and construct the noisy latent $\mathbf{z}_{\tau}$ according to the flow-matching training schedule.
The model predicts the latent velocity conditioned on both the appearance context and the geometry-control condition:
\begin{equation}
    \mathcal{L}_{\mathrm{vid}}
    =
    \mathbb{E}_{\tau}
    \left[
    \left\|
    \mathbf{v}_{\theta}
    \left(
    \mathbf{z}_{\tau}, \tau, \mathbf{C}_{a}, \mathbf{H}_{\mathrm{ctrl}}
    \right)
    -
    \mathbf{v}_{\tau}
    \right\|_{2}^{2}
    \right],
\end{equation}
where $\mathbf{v}_{\theta}$ denotes the predicted latent velocity and $\mathbf{v}_{\tau}$ denotes the flow-matching target.

\subsubsection{Inference}
At inference time, all conditions are inferred from fMRI signals.
The appearance context predictor $\mathcal{F}_{\mathrm{app}}$ predicts $\mathbf{C}_{a}$, the Geometric Predictor $\mathcal{F}_{\mathrm{geo}}$ predicts morphable 3D facial parameters, which are rendered into $\hat{\mathbf{I}}_{r}$, and the auxiliary latent predictor $\mathcal{F}_{\mathrm{aux}}$ provides $\mathbf{H}_{\mathrm{aux}}$.
The sampler denoises random video latents conditioned on $\mathbf{C}_{a}$ and $\mathbf{H}_{\mathrm{ctrl}}$, and we use classifier-free guidance with a guidance scale of 5.0 in our experiments.
The final latents are decoded by the frozen VAE.
The resulting video preserves the fMRI-inferred global appearance while following the geometry-aware facial motion decoded from brain activity.

\section{Experiments}

\begin{figure*}
  \centering
  \includegraphics[width=1.0\linewidth]{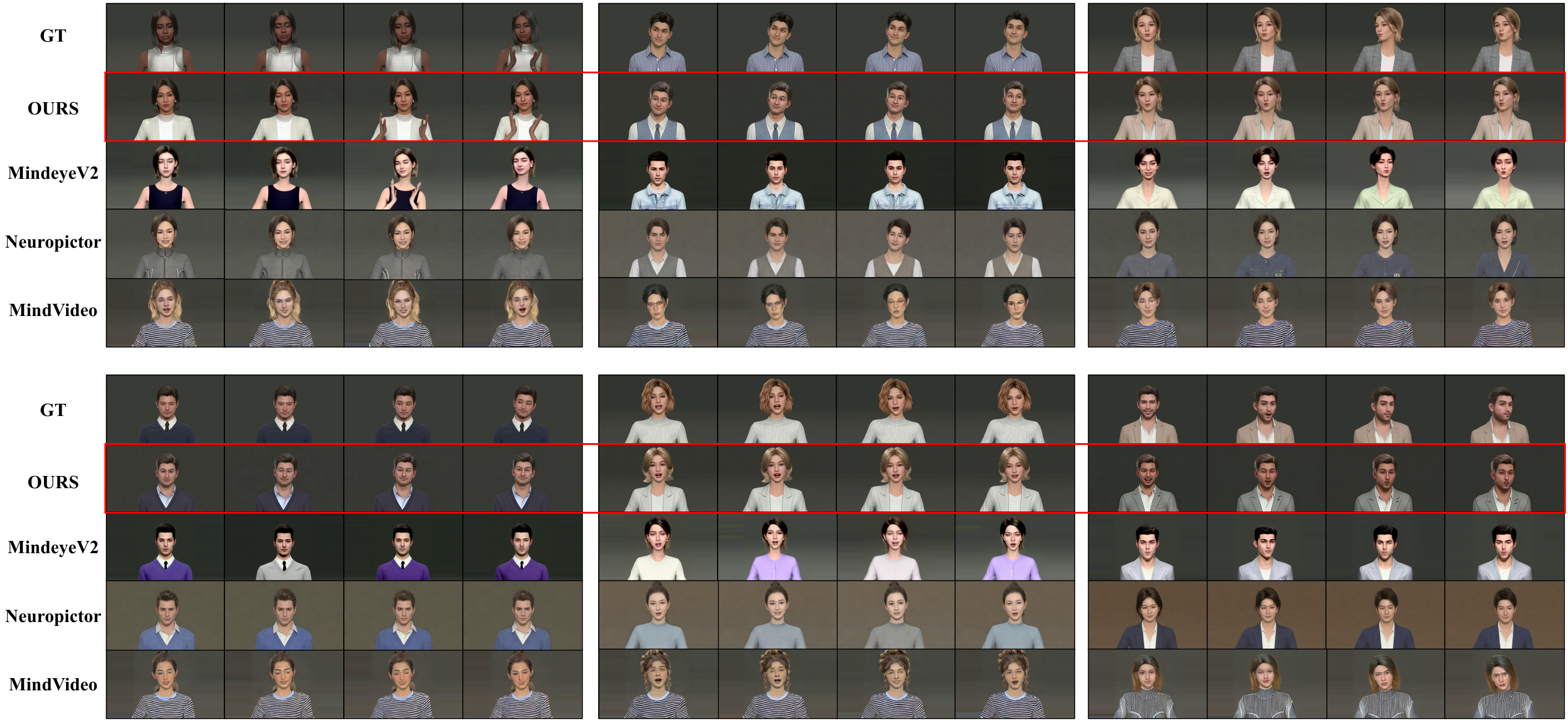}
  \caption{
 Qualitative comparison of facial video reconstruction from fMRI signals.
  For each example, four frames are shown from left to right in temporal order.
  Compared with baseline neural decoding methods, fMRI2Face better preserves identity-related appearance, pose, and expression, and produces more accurate temporal facial dynamics such as head motion and non-rigid facial changes. Zoom in for details.
  }
  \label{fig:main_results}
\end{figure*}

\begin{table*}[t]
\centering
\renewcommand{\arraystretch}{1.0}
\setlength{\tabcolsep}{3.0pt}
\caption{
Quantitative comparison with representative neural decoding methods on fMRI-Face.
All baselines are adapted and trained under the same train-test split.
The best results are shown in bold.
}
\label{tab:results}
\small
\resizebox{1.0\linewidth}{!}{
\begin{tabular}{lccccccccc}
\toprule
Method
&
PSNR $\uparrow$
&
PixCorr $\uparrow$
&
SSIM $\uparrow$
&
LPIPS $\downarrow$
&
FVD $\downarrow$
&
ID-CSIM $\uparrow$
&
LMD $\downarrow$
&
LMD-A $\downarrow$
&
FWE $\downarrow$
\\
\midrule
MindVideo~\cite{chen2023cinematic}
& 13.8726
& 0.3462
& 0.6584
& 0.3761
& 656.6294
& 0.0971
& 0.1044
& 0.1224
& 0.0288
\\
NeuroPictor~\cite{huo2024neuropictor}
& 15.5776
& 0.5184
& 0.7432
& 0.3577
& 361.3160
& 0.2281
& 0.0811
& 0.1141
& 0.0115
\\
MindEye2~\cite{scotti2024mindeye2}
& 15.7091
& 0.5808
& 0.7806
& 0.2483
& 519.1501
& 0.1915
& 0.0733
& 0.0935
& 0.0382
\\
\midrule
\textbf{fMRI2Face (OURS)}
& \textbf{18.3238}
& \textbf{0.6516}
& \textbf{0.8288}
& \textbf{0.1991}
& \textbf{82.7316}
& \textbf{0.3458}
& \textbf{0.0532}
& \textbf{0.0863}
& \textbf{0.0052}
\\
\bottomrule
\end{tabular}
}
\end{table*}

\subsection{Implementation}
We use Wan2.1-T2V-14B as the pretrained video diffusion backbone.
The video resolution is set to $480\times832$, and the Brain-derived Appearance Context consists of 64 fMRI-conditioned context tokens.
The appearance context predictor $\mathcal{F}_{\mathrm{app}}$ is trained for 100,000 iterations using AdamW with a learning rate of $1\times10^{-5}$ and a batch size of 16 on two NVIDIA H200 GPUs.
During this stage, the pretrained video diffusion model remains frozen, and only $\mathcal{F}_{\mathrm{app}}$ is optimized.
The Geometric Predictor $\mathcal{F}_{\mathrm{geo}}$ in the Morphable 3D Facial Control stream is trained for 60,000 iterations with a batch size of 8 on a single NVIDIA H200 GPU.
We use the Adam optimizer with a learning rate of $1\times10^{-4}$.
The model is supervised using the parameter-level and reprojection-level losses described in Sec.~\ref{sec:deca}.

For the final video reconstruction stage, we freeze the trained appearance context predictor and initialize the geometry-control branch from the Wan2.1-VACE-14B checkpoint.
To improve the initialization of the Auxiliary Latent Predictor $\mathcal{F}_{\mathrm{aux}}$, we first pretrain it for 8,000 iterations by aligning its predicted auxiliary latent representation with the ground-truth video VAE latent using an $\ell_2$ loss.
We then optimize the LoRA parameters of the geometry-control branch together with $\mathcal{F}_{\mathrm{aux}}$.
LoRA is applied to the attention projections, feed-forward projections, and control-bridge projections of the geometry-control branch with rank 32.
The control bridges $\mathcal{Z}_{\ell}$ are initialized to zero.
Training is performed for 10,000 iterations on two NVIDIA H200 GPUs using a batch size of 12 and a learning rate of $1\times10^{-5}$.

At test time, all conditions are inferred directly from fMRI signals and jointly injected into the video diffusion model to generate the final reconstructed facial video. We use classifier-free guidance with a guidance scale of 5.0 and 20 sampling steps during inference.

\subsection{Metrics}
We report Peak Signal-to-Noise Ratio (PSNR), Pixel Correlation (PixCorr), and Structural Similarity (SSIM)~\cite{wang2004image} to measure low-level reconstruction fidelity, and LPIPS~\cite{zhang2018perceptual} to evaluate perceptual similarity in a deep feature space. 
We use ArcFace~\cite{deng2019arcface} to compute identity cosine similarity (ID-CSIM) between reconstructed and ground-truth videos. 
To evaluate facial geometry, we report Landmark Distance (LMD) and its aligned-crop version (LMD-A), computed as the mean Euclidean distance between reconstructed and ground-truth 2D facial landmarks in the original image coordinate space and the aligned face-crop coordinate space, respectively.
We further use Fréchet Video Distance (FVD)~\cite{unterthiner2018towards} to measure overall video quality and temporal realism, and Flow Warping Error (FWE) to evaluate motion consistency by comparing reconstructed frame transitions under optical-flow-based warping from the ground-truth video.
Additional details on the evaluation metrics are provided in the supplementary material.

\begin{table*}[t]
\centering
\renewcommand{\arraystretch}{1.0}
\setlength{\tabcolsep}{2.7pt}
\caption{Ablation study of key components in fMRI2Face. ``--'' indicates that face detection failed and the corresponding face-specific metric could not be computed.}
\label{tab:ablation}
\small
\resizebox{1.0\linewidth}{!}{
\begin{tabular}{lcccccccccc}
\toprule
Variant
&
Tokens
&
PSNR $\uparrow$
&
PixCorr $\uparrow$
&
SSIM $\uparrow$
&
LPIPS $\downarrow$
&
FVD $\downarrow$
&
ID-CSIM $\uparrow$
&
LMD $\downarrow$
&
LMD-A $\downarrow$
&
FWE $\downarrow$
\\
\midrule
Appearance only
& 1
& 5.5453 & -0.1774 & 0.4952 & 0.5673 & 4124.9912 & -- & -- & -- & 0.0109
\\
Appearance only
& 16
& 14.5522 & 0.5836 & 0.7356 & 0.2579 & 265.4814 & 0.2559 & 0.0670 & 0.0902 & 0.0063
\\
Appearance only
& 64
& 16.7073 & 0.6328 & 0.8165 & 0.2047 & 153.7794 & 0.3123 & 0.0589 & 0.0919 & 0.0057
\\
\midrule
+ 3D Control
& 64
& 17.4465 & 0.6281 & 0.8257 & 0.1948 & 96.2231 & 0.3643 & 0.0485 & 0.0844 & 0.0054
\\
+ Aux. Latent
& 64
& 18.6546 & 0.6754 & 0.8348 & 0.1892 & 84.5020 & 0.3608 & 0.0485 & 0.0840 & \textbf{0.0052} 
\\
+ Mask \textbf{(Full)}
& 64
& \textbf{18.7449} & \textbf{0.6822} & \textbf{0.8353} & \textbf{0.1885} & \textbf{78.6234} & \textbf{0.3682} & \textbf{0.0484} & \textbf{0.0833} & \textbf{0.0052}
\\
\bottomrule
\end{tabular}
}
\end{table*}

\subsection{Main Results}
\subsubsection{Quantitative Results}

We compare fMRI2Face with representative neural decoding methods, including MindVideo~\cite{chen2023cinematic}, NeuroPictor~\cite{huo2024neuropictor}, and MindEye2~\cite{scotti2024mindeye2}.
For a fair comparison, all baselines are adapted and trained on the proposed fMRI-Face dataset under the same train-test split.

As shown in Tab.~\ref{tab:results}, fMRI2Face achieves the best performance across all evaluation metrics.
\textbf{(1)} Compared with the best baseline on each metric, our method improves PSNR from 15.7091 to 18.3238 and PixCorr from 0.5808 to 0.6516, indicating more faithful low-level reconstruction.
It also achieves the highest SSIM and the lowest LPIPS, suggesting better structural preservation and perceptual similarity.
\textbf{(2)} For video-level evaluation, fMRI2Face substantially reduces FVD from 361.3160 to 82.7316, demonstrating improved visual realism and temporal coherence. 
It also lowers FWE from 0.0115 to 0.0052, suggesting that the reconstructed videos better follow the motion patterns of the ground-truth facial videos.
\textbf{(3)} For face-specific evaluation, fMRI2Face also shows consistent advantages.
It improves ID-CSIM from 0.2281 to 0.3458, indicating stronger preservation of identity-related facial appearance.
In addition, our method obtains the lowest LMD and LMD-A, showing more accurate recovery of facial geometry.
Overall, these results demonstrate that fMRI2Face improves not only general reconstruction fidelity but also the key properties of dynamic face perception, including identity, facial structure, and motion consistency.

\subsubsection{Qualitative Results}
Fig.~\ref{fig:main_results} provides qualitative comparisons between our fMRI2Face and baseline methods.
The baseline reconstructions generally recover coarse facial layouts but often exhibit blurred details, inaccurate expressions, or weak temporal correspondence across frames.
In contrast, fMRI2Face produces more identity-consistent facial videos that better match the ground truth in facial appearance, skin tone, pose, and expression.
In particular, our method more faithfully captures dynamic facial changes such as head orientation and mouth movement, which are difficult to infer from fMRI signals.
These qualitative results further validate the effectiveness of the proposed framework for high-fidelity facial video reconstruction from brain activity.

\begin{figure*}[t]
    \centering
    \begin{subfigure}[t]{0.6\linewidth}
        \centering
        \includegraphics[width=0.9\linewidth]{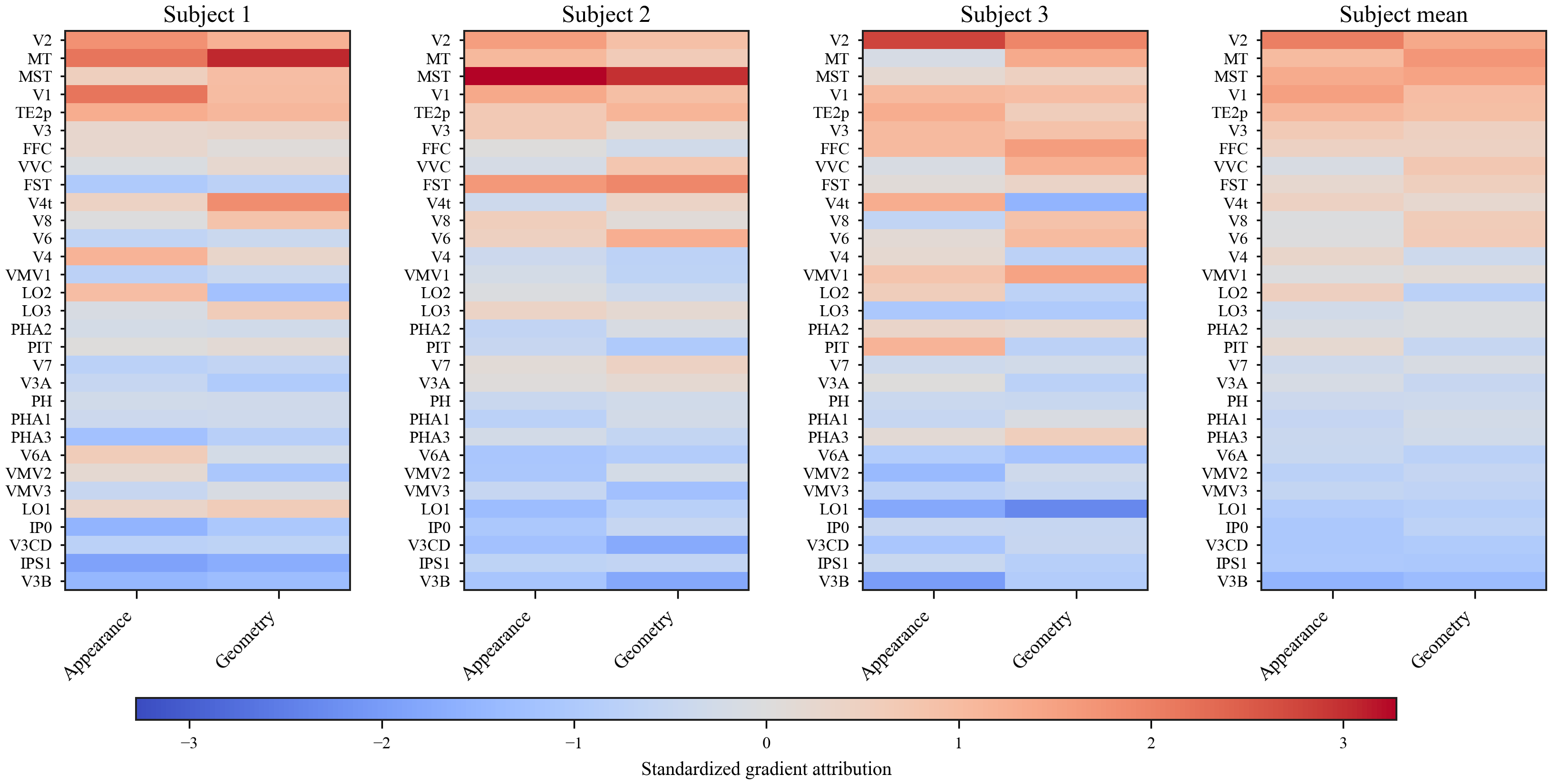}
        \caption{Subject-specific and subject-mean ROI attribution.}
        \label{fig:roi_gradient_heatmaps}
    \end{subfigure}
    \hfill
    \begin{subfigure}[t]{0.32\linewidth}
        \centering
        \includegraphics[width=0.9\linewidth]{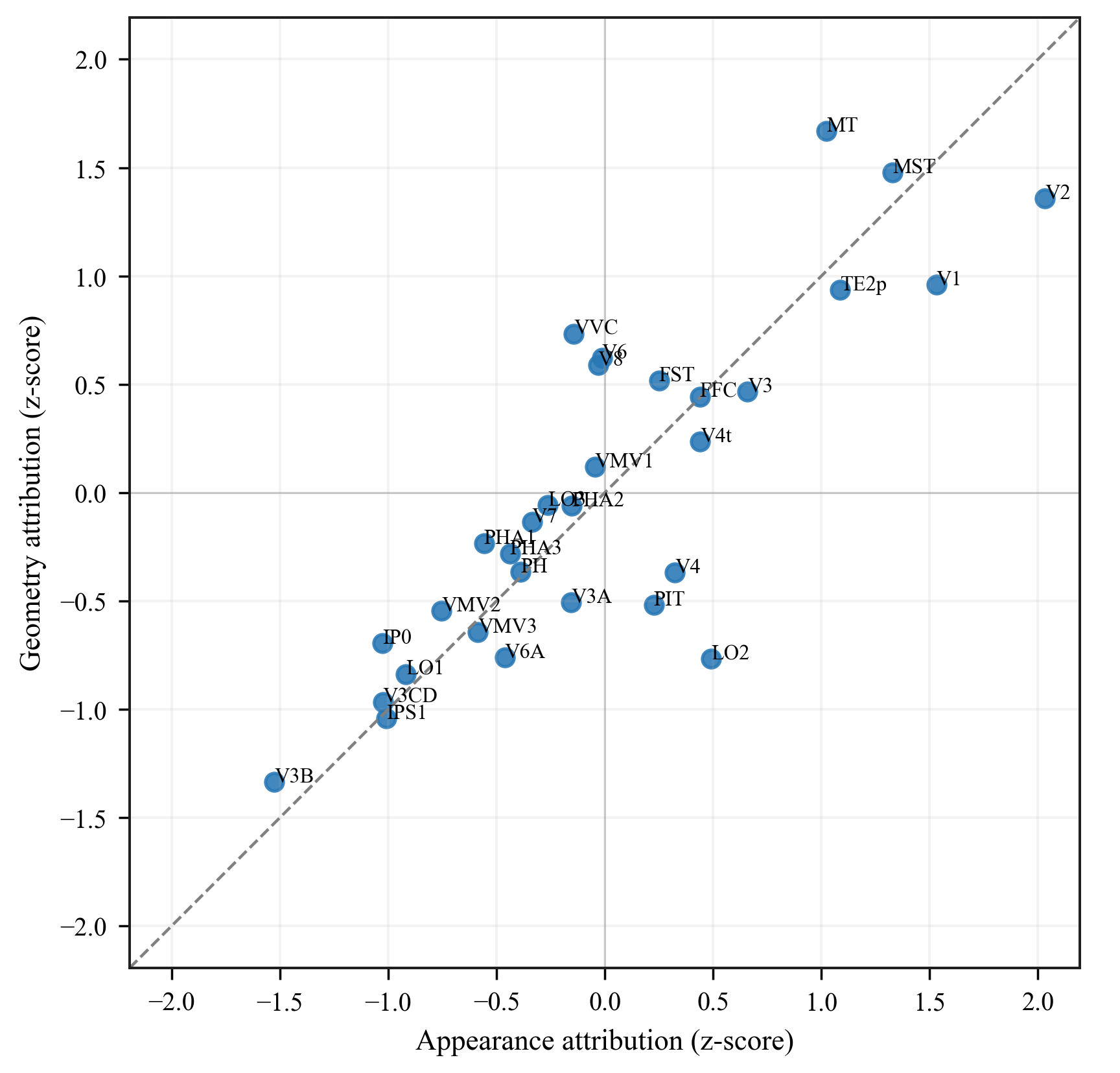}
        \caption{Appearance--Geometry comparison.}
        \label{fig:roi_gradient_scatter}
    \end{subfigure}
    \caption{
    Gradient-based ROI attribution for the Brain-derived Appearance Context and Morphable 3D Facial Control streams. (a) The first three heatmaps show gradient attribution standardized within each subject, and the fourth shows the mean of these scores across subjects. (b) The scatter plot compares subject-mean Appearance and Geometry attribution scores, with the dashed line indicating equal relative attribution to both streams.
    }
    \label{fig:roi_gradient}
\end{figure*}

\begin{figure*}[t]
    \centering
    \includegraphics[width=0.75\linewidth]{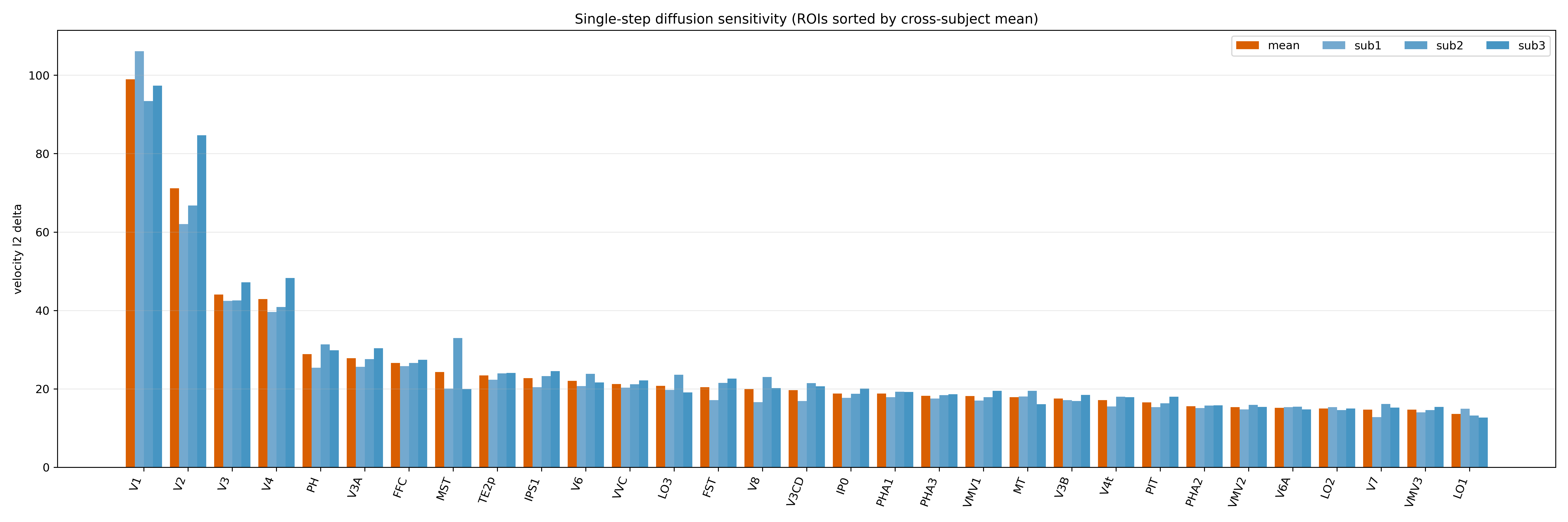}
    \caption{
    Single-step diffusion sensitivity to ROI masking. 
    For each ROI, the grouped bars show the mean across subjects and the scores of subjects 1--3. 
    Larger values indicate greater changes in the velocity prediction after masking ROI signals.
    }
    \label{fig:roi_diffusion}
\end{figure*}

\subsection{Ablation Study}

We conduct ablation experiments on subject 1 to evaluate the key components of fMRI2Face.
Results are shown in Tab.~\ref{tab:ablation}.
\textbf{(1)} We start from an appearance-only setting, where only the Brain-derived Appearance Context is used to condition the video diffusion model.
We analyze the effect of the number of context tokens.
With only one token, the model often fails to generate recognizable human faces, making several face-specific metrics invalid.
Increasing the number of tokens substantially improves reconstruction quality, and using 64 tokens consistently outperforms 16 tokens across most metrics.
This indicates that a richer fMRI-conditioned context is important for representing global facial appearance.
\textbf{(2)} Adding Morphable 3D Facial Control (``+ 3D Control'') improves most metrics, especially those related to facial geometry and video-level consistency.
LMD decreases from 0.0589 to 0.0485, and LMD-A decreases from 0.0919 to 0.0844, showing that the rendered 3D facial guidance provides effective constraints for facial structure, pose, and expression.
The improvement in FVD and ID-CSIM further suggests that geometry-aware facial motion guidance benefits both temporal realism and identity-related appearance preservation.
\textbf{(3)} Introducing the auxiliary latent further improves low-level reconstruction fidelity, as reflected by higher PSNR, PixCorr, and SSIM. This suggests that the auxiliary latent helps recover more complete visual information. 
We observe a slight decrease in ID-CSIM after adding it, which may be caused by additional latent context introducing minor appearance variations.
\textbf{(4)} Finally, adding the binary mask yields the best overall performance. By indicating the valid rendered facial-control region, the mask helps the diffusion model use the morphable facial guidance more effectively while balancing it with the auxiliary latent context. 
It also recovers the slight ID-CSIM drop caused by the auxiliary latent, leading to the strongest identity preservation. 
These results support the full fMRI2Face design.

\subsection{ROI-Based Model Analysis}
\label{sec:roi_analysis}

To better understand how fMRI2Face uses neural signals from different visual regions, we conduct ROI-based analyses at two levels.
First, we examine gradient attribution for the two fMRI-derived conditioning streams.
Second, we assess downstream diffusion sensitivity by masking individual ROIs and measuring changes in predicted diffusion velocity.
These analyses characterize how different visual regions contribute to the trained decoding pipeline.

\subsubsection{Gradient Attribution}
We first analyze which visual ROIs contribute to the Brain-derived Appearance Context and Morphable 3D Facial Control streams using gradient-times-input attribution.
For sample $i$, vertex $v$, and stream $b \in \{\mathrm{app}, \mathrm{geo}\}$, the attribution is computed as
\begin{equation}
A^{b}_{i,v}
=
\sum_{t}
\left|
x_{i,t,v}
\frac{\partial \Phi^{b}_{i}}{\partial x_{i,t,v}}
\right|,
\end{equation}
where $x_{i,t,v}$ denotes the fMRI response at time $t$ and voxel $v$, and $\Phi^{b}_{i}$ is the squared-magnitude target of the corresponding stream, i.e., the appearance context for $b=\mathrm{app}$ or the predicted morphable 3D facial parameters for $b=\mathrm{geo}$.
We average voxel-level attributions over test samples and within each ROI.
To compare across subjects, ROI-wise scores are independently z-scored within each subject and stream, and then averaged across subjects.

As shown in Fig.~\ref{fig:roi_gradient}, both streams rely strongly on early visual and face- and motion-related regions, but with different preferences.
For the Brain-derived Appearance Context stream, the highest-attribution ROIs include V2, V1, MT, MST, and TE2p, suggesting strong contributions from early visual areas.
For the Morphable 3D Facial Control stream, MT, MST, V2, V1, and TE2p form the most influential set, indicating a stronger emphasis on motion- and geometry-related information.
The scatter plot compares the relative attribution of the two streams.
Appearance-leaning regions, which show higher attribution for appearance than geometry, include V2, V1, TE2p, V3, V4t, V4, PIT, and LO2.
Geometry-leaning regions include MT, MST, V6, V8, FST, VVC, and VMV1, with MT and MST showing the strongest geometry dominance.
Several regions, such as FFC, PH, V7, PHA2, V3CD, IPS1, and LO1, lie close to the diagonal, indicating more balanced contributions to the two streams.

\subsubsection{Diffusion Sensitivity}

We further evaluate how strongly each ROI affects the downstream video diffusion pathway.
For each test sample, we first compute the diffusion velocity prediction $\mathbf{v}_{i}$ using the full fMRI input.
We then mask the vertices of ROI $r$ and recompute the velocity prediction $\mathbf{v}_{i,r}^{\mathrm{mask}}$ at the same diffusion time.
The diffusion sensitivity score for ROI $r$ is defined as
\begin{equation}
D_r
=
\frac{1}{N}
\sum_{i=1}^{N}
\left\|
\mathbf{v}_{i}
-
\mathbf{v}_{i,r}^{\mathrm{mask}}
\right\|_2 .
\end{equation}
We compute this score using a single diffusion step and average the results across test samples and subjects.

As shown in Fig.~\ref{fig:roi_diffusion}, the diffusion pathway is most sensitive to early visual regions.
V1 has the largest effect, followed by V2, V3, and V4, while PH, V3A, FFC, and MST form the next tier.
This indicates that the final video generation process is strongly affected by low- and mid-level visual representations, which provide important spatial and appearance-related information for facial video reconstruction.
The subject-specific bars show a broadly consistent ranking across participants, suggesting that the learned diffusion sensitivity profile is relatively stable across subjects.

\section{Conclusion}

In this work, we introduced \textbf{fMRI-Face}, the first large-scale fMRI dataset paired with controllable, full-HD digital human facial videos, and \textbf{fMRI2Face}, a geometry-guided neural video decoding framework for reconstructing dynamic faces from brain activity.
By providing background-free and parameterized facial stimuli with controlled identity, expression, and head pose, fMRI-Face offers a structured platform for studying neural representations underlying dynamic face perception.
Building on this dataset, fMRI2Face integrates Brain-derived Appearance Context, Morphable 3D Facial Control, and Neural-Controlled Video Diffusion to recover both global facial appearance and geometry-aware facial dynamics.
Extensive experiments demonstrate that this design improves identity preservation, facial structure recovery, and motion consistency over representative neural decoding baselines.
Together, fMRI-Face and fMRI2Face establish a new benchmark for fMRI-based digital human reconstruction and provide a step toward neural decoding systems that connect brain activity with controllable, dynamic facial video generation.

\section*{Ethics statement}
This study was approved by the Ethics Committee of Fudan University. All participants were informed of the purpose and procedures of the study and provided written informed consent before participation. 

\bibliographystyle{cas-model2-names}

\bibliography{cas-refs}

\clearpage
\noindent \textbf{\Large Supplementary Material}
\vspace{0.1in}
\setcounter{section}{0}

\section{Dataset}

\subsection{More Details}
We provide additional details on the digital human composition and annotations in the fMRI-Face dataset.
In total, fMRI-Face contains 2,174 digital human videos, including 1,193 female and 981 male samples.
Among them, 1,694 videos feature artistic digital humans, which are high-quality characters created using commercial digital-human authoring tools and further refined by professional artists.
These characters include diverse facial textures, hairstyles, and garments, covering 13 clothing categories with rich variations in color and style.
The remaining 480 videos consist of parametric digital humans generated by interpolating facial geometry meshes over a shared base-mesh library.
Compared with artistic digital humans, these digital humans exhibit more uniform visual traits, including four fixed hairstyles, two skin textures, and a set of predefined outfits assigned without additional customization.
The gender distribution across the two types of digital humans is summarized in Tab.~\ref{tab:gender_distribution}.

To support future research from multiple perspectives, each digital human video in fMRI-Face is annotated with a canonical portrait, clothing category, hairstyle category, motion configuration, and PCA-based facial mesh parameters.
We describe these annotations in detail below.

\begin{figure*}[t]
  \begin{centering}
\includegraphics[width=0.95\linewidth]{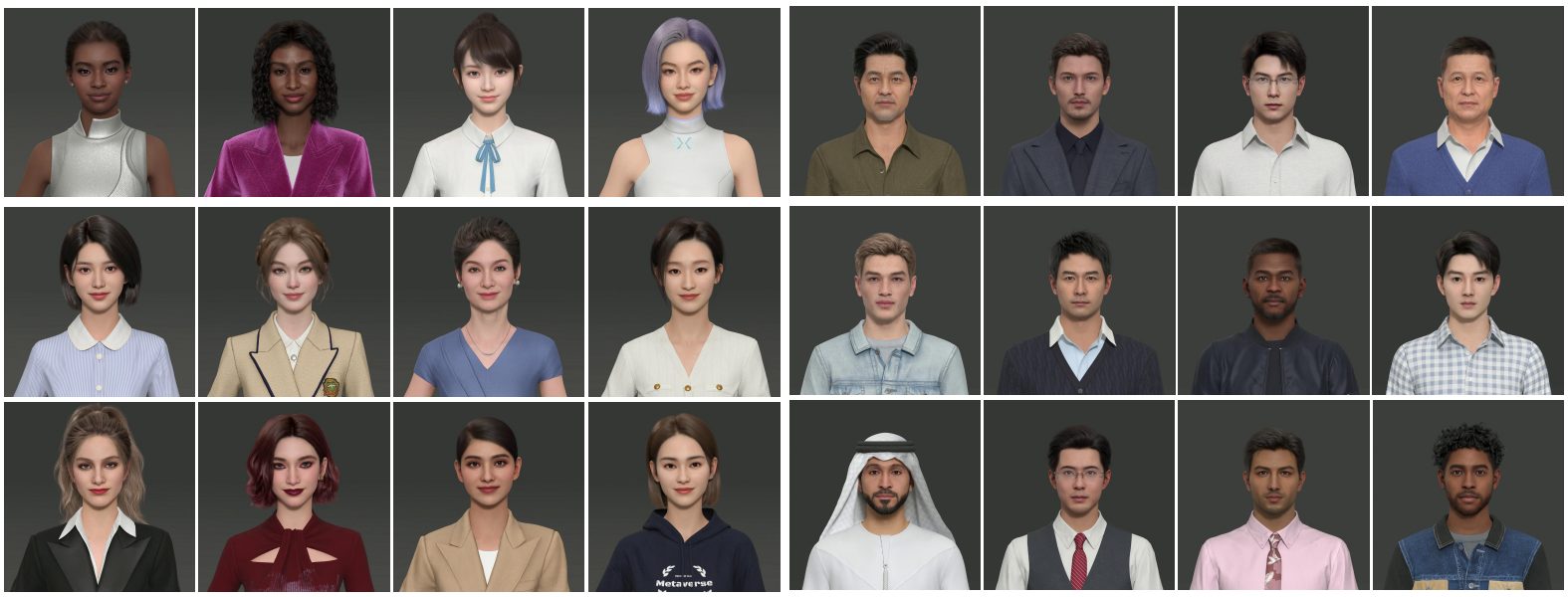}
\par\end{centering}
  \caption{Visualization of canonical portraits.}
  \label{fig:canonical_portrait}
\end{figure*}

\noindent
\textbf{Canonical Portrait.}
For each digital human video, we provide a corresponding canonical portrait, rendered under neutral expression, frontal pose, and uniform lighting.
This portrait, denoted as $cp$, offers a clean and consistent view of the digital human's facial appearance, as shown in Fig.~\ref{fig:canonical_portrait}.
Because all portraits follow the same rendering protocol, they facilitate identity inspection, region-specific feature extraction (e.g., eyes, mouth, jawline) and facial-region mask generation.

\begin{figure}
  \begin{centering}
\includegraphics[width=1.0\linewidth]{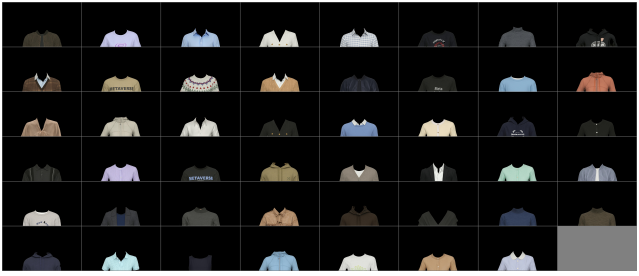}
\par\end{centering}
  \caption{Sample visualization of the 47 fine-grained clothing items used for Parametric Digital Humans.}
  \label{fig:cloth_grid}
\end{figure}

\noindent
\textbf{Clothing.}
We provide two levels of clothing annotations.
The first is a coarse-level label that assigns each outfit to one of 13 high-level categories, as summarized in Tab.~\ref{tab:cloth_annot}.
Each category contains multiple clothing items with variations in style and color.
The second is a fine-grained instance-level annotation covering 47 specific clothing items.
This instance-level annotation is available only for parametric digital humans, whose clothing is selected from a predefined item set without additional customization.
In contrast, artistic digital humans use artist-edited clothing with diverse colors and styles, making instance-level labeling infeasible.
A visualization of the 47 clothing items is shown in Fig.~\ref{fig:cloth_grid}.

\noindent
\textbf{Hair.}
Hair is categorized into four types: buzz cut and side part for male digital humans, and short loose hair and bun for female digital humans. These labels apply only to the parametric digital humans, whose hairstyles are fixed and exhibit no variation in color or fine-grained styling.

\noindent
{\bf Motion.} 
We characterize motion in fMRI-Face using both controller-level annotations and motion-configuration labels.
First, we provide low-level controller parameters that describe frame-by-frame facial dynamics.
These parameters govern localized deformations across facial regions, including 62 controls for mouth movement, 14 for eyelid motion, 3 for gaze direction, and 7 for eyebrow movement.
Together, they provide a continuous and fine-grained representation of expression changes over time.
Second, each digital human is animated using a predefined 45-second facial performance sequence. For stimulus construction, we generate video clips by randomly sampling 8-second segments from this performance using integer-second starting points. This procedure yields 16 distinct motion configurations for female digital humans (covering approximately 23 seconds of the full animation) and 15 for male digital humans (covering 22 seconds), resulting in a total of 31 unique motion types in the dataset.

\begin{figure}
  \begin{centering}
\includegraphics[width=1.0\linewidth]{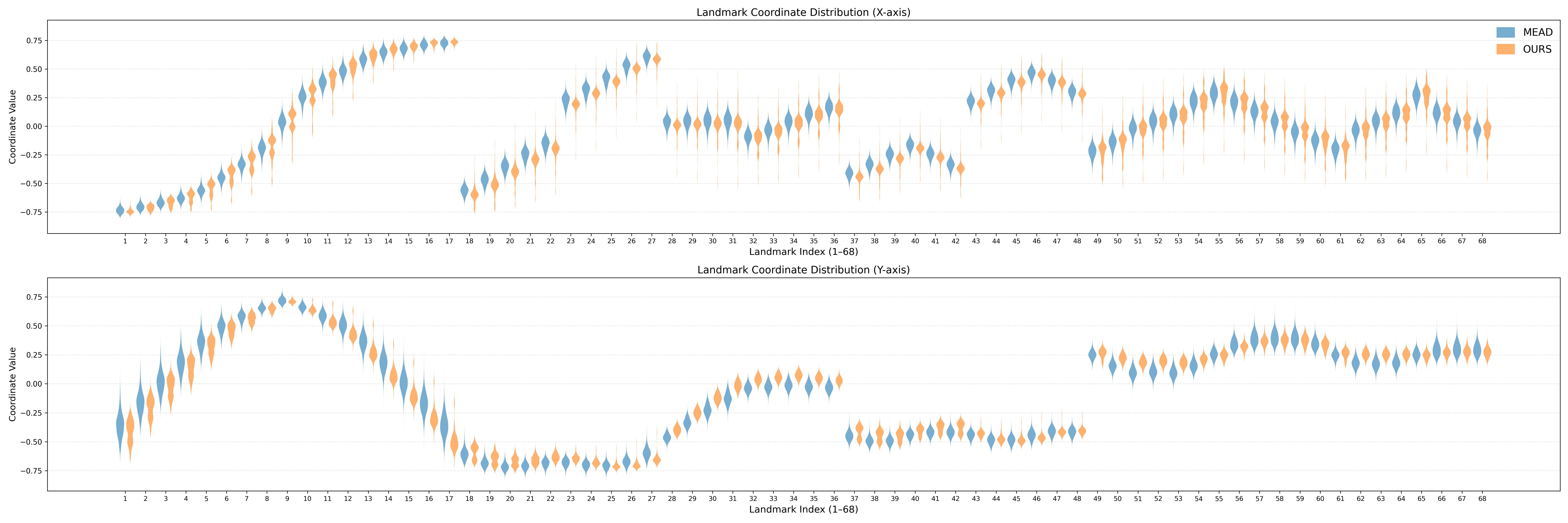}
\par\end{centering}
  \caption{
  Landmark-coordinate distribution comparison between fMRI-Face and MEAD.
  We extract 68 2D facial landmarks from both datasets and visualize their normalized coordinate distributions using violin plots.
The upper and lower panels show the $x$- and $y$-coordinate distributions, respectively.
Blue denotes MEAD and orange denotes fMRI-Face.
  }
  \label{fig:motion}
\end{figure}

To further examine the naturalness and variability of the facial motion in fMRI-Face, we compare it with MEAD~\cite{wang2020mead}, a real-human facial video dataset.
Specifically, we extract 68 2D facial landmarks from both datasets and analyze the coordinate distributions of each landmark.
As shown in Fig.~\ref{fig:motion}, we visualize the landmark-wise distributions using violin plots, with the upper panel showing the normalized $x$-coordinate distributions and the lower panel showing the normalized $y$-coordinate distributions.
Blue denotes MEAD and orange denotes fMRI-Face.
Although fMRI-Face is generated from controlled digital human animations rather than real video recordings, its landmark-coordinate distributions remain broadly similar to those of MEAD.
While local differences can be observed for some landmarks, the overall distribution patterns and coordinate ranges are largely consistent across the two datasets.
Quantitatively, 90.3\% of the MEAD landmark value range overlaps with that of fMRI-Face, further indicating that our dataset covers a similar range of facial motion.

\begin{table}
\centering
\caption{Gender distribution in the fMRI-Face dataset. }
\label{tab:gender_distribution}
\begin{tabular}{lcc}
\toprule
\textbf{Type} & \textbf{Female} & \textbf{Male} \\
\midrule
Artistic Digital Humans     & 992   & 702   \\
Parametric Digital Humans   & 201   & 279   \\
\midrule
All                  & 1193  & 981   \\
\bottomrule
\end{tabular}
\end{table}

\begin{table}
\centering
\caption{Coarse clothing category distribution in fMRI-Face.
The table reports the number of videos for each clothing category.}
\label{tab:cloth_annot}
\resizebox{\linewidth}{!}
{
\begin{tabular}{lcccccccc}
\toprule
Clothing & Suit & Shirt & Sweater & Hoodie & T-shirt \\
\midrule
Number & 616  & 390   & 337     & 251    & 174 \\
\midrule
Clothing & Dress & Jacket & Bodysuit & Jumpsuit & Undershirt \\
\midrule
Number & 169   & 138    & 28       & 25       & 21      \\
\midrule
Clothing  & Vest & Coat & Overcoat \\
\midrule
Number  & 11   & 10   & 4         \\
\bottomrule
\end{tabular}
}
\end{table}

\noindent
\textbf{Facial Mesh PCA.}
We represent the underlying 3D facial geometry of each digital human using PCA-based shape parameters derived from a shared 3D facial base corpus.
For artistic digital humans, we define 136 distinct face shapes.
Each shape is fixed, while variations in surface-level attributes, including skin texture, eyebrows, eyelashes, eye color, hairstyle, and clothing, produce multiple visually diverse videos.
For parametric digital humans, we generate 480 unique shapes by interpolating from the same standard template.
This PCA representation captures the core structural variation across digital humans.

\subsection{Discussion: Why Digital Humans?}

In designing fMRI-Face, we deliberately chose high-fidelity digital human videos rather than recordings of real people.
This choice is motivated by both methodological considerations and the practical constraints of fMRI data collection.

First, real-face videos inevitably couple facial appearance and motion with uncontrolled environmental factors, such as changing backgrounds, illumination, shadows, camera viewpoints, and recording artifacts.
These factors introduce stimulus variations that are orthogonal to the core scientific question: how the brain encodes facial identity, expression, and dynamic facial motion.
In contrast, our digital human stimuli are rendered under a unified environment with fixed lighting and background-free composition.
This controlled design removes many nuisance variables and ensures that perceptual variations arise primarily from the face itself, including identity, expression, and pose.
As a result, the fMRI responses can be more directly associated with facial properties, reducing the sample complexity required to learn fMRI-to-face mappings.
Achieving comparable control and coverage with real videos would require substantially larger fMRI acquisitions, leading to prohibitive scanning time and cost.
Together with the larger scale of fMRI-Face compared with prior face-focused datasets~\cite{vanrullen2019reconstructing,dado2022hyperrealistic,chen2024fmri}, this controlled stimulus design provides a practical and effective foundation for high-quality facial video decoding.

Second, recent advances in digital human technology have substantially narrowed the realism gap between rendered and real faces.
Production pipelines used in virtual anchors, streaming media, and interactive entertainment can generate facial textures, expressions, and motions that closely approximate natural human appearance and behavior.
The stimuli in fMRI-Face follow this production-level workflow, combining artistic digital humans, physically based facial rigs, and high-quality expressive animation.
This enables us to use digital human videos as controlled yet realistic stimuli for studying neural representations of facial appearance and dynamics, while maintaining a level of experimental control that real-world footage cannot provide.

Finally, scientific progress often requires balancing ideal stimuli with practical constraints. 
Current fMRI technology demands long acquisition times, substantial participant effort, and is limited in temporal resolution. Within these constraints, high-fidelity digital humans provide an effective bridge: they are realistic enough to engage the brain’s natural face-processing pathways, yet controlled enough to support systematic investigation of identity, expression, and motion encoding. 
We view the fMRI-Face dataset as a step toward future studies with increasingly naturalistic facial stimuli, while providing an efficient, scalable, and interpretable benchmark for dynamic face decoding today.

\section{More Architecture Details}

\subsection{Brain-derived Appearance Context}
\label{sec:supp_app_arch}

We provide the implementation details of the Appearance Context Predictor $\mathcal{F}_{\mathrm{app}}$.
Given an fMRI input window $\mathbf{F}\in\mathbb{R}^{T\times d}$ with $T=5$ and $d=8{,}921$, we first project each fMRI measurement into the context dimension of the video diffusion transformer.
Sinusoidal temporal positional encodings are added to preserve the temporal order of the fMRI window, and a learnable class token is prepended to the sequence.

The resulting fMRI tokens are used as memory tokens for a Transformer decoder.
We use $N_a=64$ learnable query tokens, each with dimension $D_c=4096$, to extract a fixed-length set of appearance context tokens:
\begin{equation}
    \mathbf{C}_{a}
    =
    \mathrm{LN}
    \left(
    \mathrm{Dec}_{\mathrm{app}}
    (
    \mathbf{Q}_{a},
    \mathbf{M}_{\mathrm{app}}
    )
    \right),
\end{equation}
where $\mathbf{Q}_{a}\in\mathbb{R}^{64\times4096}$ denotes the learnable appearance queries, $\mathbf{M}_{\mathrm{app}}$ denotes the projected fMRI memory tokens, and $\mathrm{Dec}_{\mathrm{app}}$ is a one-layer Transformer decoder.
The output $\mathbf{C}_{a}\in\mathbb{R}^{64\times4096}$ is directly fed into the original context-conditioning interface of the pretrained video diffusion transformer, replacing the prompt-derived text context.

\subsection{Morphable 3D Facial Control}
\label{sec:supp_geo_arch}

We provide additional implementation details of the Geometric Predictor $\mathcal{F}_{\mathrm{geo}}$.
The predictor maps an fMRI input window $\mathbf{F}$ into a 24-frame sequence of DECA-based facial control parameters.
Each fMRI measurement is first projected to a 512-dimensional hidden space, followed by sinusoidal temporal positional encoding.
A learnable class token is prepended to the sequence, and the resulting tokens are processed by a two-layer Transformer encoder with 8 attention heads.

From the class-token feature, the model predicts a first-frame reference state, including a 236-dimensional DECA parameter vector, a 128-dimensional detail code, and a 6-dimensional image transformation vector.
The 6-dimensional transformation represents the affine transformation matrix.
To model temporal facial dynamics, the same class-token feature is further mapped to $K-1=23$ residual states.
Together with the first-frame reference state, these residuals form a $K=24$ frame control sequence:
\begin{equation}
    (
    \hat{\mathbf{P}},
    \hat{\mathbf{D}},
    \hat{\mathbf{T}}
    )
    =
    \mathcal{F}_{\mathrm{geo}}(\mathbf{F}),
\end{equation}
where
$\hat{\mathbf{P}}\in\mathbb{R}^{24\times236}$ denotes the DECA parameter sequence,
$\hat{\mathbf{D}}\in\mathbb{R}^{24\times128}$ denotes the detail-code sequence, and
$\hat{\mathbf{T}}\in\mathbb{R}^{24\times6}$ denotes the image-transformation sequence.

The residual prediction is applied only to temporally varying components, including expression, pose, detail, and image transformation.
Stable attributes such as shape, albedo, and illumination are copied from the first-frame reference prediction across the clip.
This design anchors the identity-related facial structure while allowing the model to focus its temporal capacity on expression and pose changes.

During training, ground-truth DECA parameters are extracted from target video frames using the frozen DECA encoder.
We apply min--max normalization to the DECA parameter vector, detail code, and image transformation before supervision, and denormalize the predicted parameters before rendering.
The denormalized parameters are decoded and rendered by the frozen DECA decoder and renderer to obtain the rendered facial guidance sequence:
\begin{equation}
    \hat{\mathbf{I}}_{r}
    =
    \mathcal{R}_{\mathrm{DECA}}
    \left(
    \mathcal{D}_{\mathrm{DECA}}
    (
    \hat{\mathbf{P}},
    \hat{\mathbf{D}}
    ),
    \hat{\mathbf{T}}
    \right).
\end{equation}
This rendered sequence is used as explicit geometry-aware facial control for the video diffusion model.

\subsection{Neural-Controlled Video Diffusion}
\label{sec:supp_diff_arch}

We provide additional details of the Neural-Controlled Video Diffusion module.
The module is built on Wan2.1-T2V-14B as the main video diffusion backbone, while the geometry-control branch is initialized from Wan2.1-VACE-14B.
The model operates on 24-frame clips at a spatial resolution of $480\times832$.

The Brain-derived Appearance Context $\mathbf{C}_{a}$ is injected through the original context-conditioning interface of the pretrained denoising transformer.
Thus, the prompt-derived text context is replaced by fMRI-conditioned appearance tokens, while the pretrained context projection and attention interface are preserved.

For geometry control, the rendered Morphable 3D Facial Control sequence $\hat{\mathbf{I}}_{r}$ is encoded by the frozen Wan VAE to obtain the rendered-control latent:
\begin{equation}
    \mathbf{H}_{r}
    =
    \mathcal{E}_{\mathrm{VAE}}(\hat{\mathbf{I}}_{r}).
\end{equation}
We also derive a binary mask sequence $\mathbf{S}_{r}$ from the rendered facial region.
Since the rendered DECA guidance mainly covers the face region, we introduce an auxiliary latent predictor $\mathcal{F}_{\mathrm{aux}}$ to provide additional fMRI-derived latent context:
\begin{equation}
    \mathbf{H}_{\mathrm{aux}}
    =
    \mathcal{F}_{\mathrm{aux}}(\mathbf{F}).
\end{equation}
The final geometry-control condition is constructed by concatenating the rendered-control latent, auxiliary latent, and downsampled mask:
\begin{equation}
    \mathbf{H}_{\mathrm{ctrl}}
    =
    \mathrm{Concat}
    \left(
    \mathbf{H}_{r},
    \mathbf{H}_{\mathrm{aux}},
    \mathrm{Down}(\mathbf{S}_{r})
    \right).
\end{equation}

The geometry-control branch processes $\mathbf{H}_{\mathrm{ctrl}}$ and injects control features into the main denoising transformer through zero-initialized residual bridges.
At the beginning of training, these bridges introduce no perturbation to the pretrained generator.
During adaptation, the control branch learns to incorporate fMRI-derived facial geometry and motion cues.
We train this branch using LoRA with rank 32, while keeping the pretrained denoising backbone, video VAE, and Appearance Context Predictor frozen.

\section{Experiments}

\subsection{More Implementation Details}

We provide additional training details for the three stages of fMRI2Face.
All video diffusion experiments are conducted at a resolution of $480\times832$ with 24 frames.

\noindent
\textbf{Brain-derived Appearance Context.}
We use Wan2.1-T2V-14B as the pretrained video diffusion backbone.
The Appearance Context Predictor $\mathcal{F}_{\mathrm{app}}$ predicts 64 context tokens with dimension 4096.
It is trained for 100{,}000 iterations using AdamW with a learning rate of $1\times10^{-5}$ and a batch size of 16 on two NVIDIA H200 GPUs.
During this stage, the pretrained video diffusion model remains frozen, and only $\mathcal{F}_{\mathrm{app}}$ is optimized.

\noindent
\textbf{Morphable 3D Facial Control.}
The Geometric Predictor $\mathcal{F}_{\mathrm{geo}}$ is trained for 60{,}000 iterations with a batch size of 8 on a single NVIDIA H200 GPU.
We use the Adam optimizer with a learning rate of $1\times10^{-4}$.
The model is supervised by the parameter-level, temporal-difference, landmark reprojection, lip, and eye losses described in the main paper.
During training, DECA parameters extracted from the target video frames are used as supervision, and the DECA encoder, decoder, and renderer remain frozen.
When computing the landmark reprojection loss $\mathcal{L}_{\text{lmk}}$, we use two complementary terms.  
The first is computed on the cropped face input, which follows the standard DECA preprocessing.  
The second is computed on the original full image before cropping, which requires estimating an additional transformation parameter $\boldsymbol{t}_{\text{form}}$ that maps coordinates from the full frame to the DECA crop region.  
These two terms are summed to obtain the final $\mathcal{L}_{\text{lmk}}$.  
For the overall loss $\mathcal{L}_{\text{total}}$, we use the weighting coefficients  
$\lambda_{\text{param}}=1.0$,  
$\lambda_{\text{lmk}}=1.0$,  
$\lambda_{\text{lip}}=0.5$,  
$\lambda_{\text{eye}}=1.0$,
$\lambda_{\Delta\mathrm{param}}=1.0$,
and $\lambda_{\Delta\mathrm{lmk}}=1.0$.

\noindent
\textbf{Neural-Controlled Video Diffusion.}
For the final video reconstruction stage, we freeze the trained Appearance Context Predictor and initialize the geometry-control branch from the Wan2.1-VACE-14B checkpoint.
Before diffusion-stage training, we pretrain the Auxiliary Latent Predictor $\mathcal{F}_{\mathrm{aux}}$ for 8{,}000 iterations by aligning its predicted auxiliary latent representation with the ground-truth video VAE latent using an $\ell_2$ loss.
We then jointly optimize $\mathcal{F}_{\mathrm{aux}}$ and the LoRA parameters of the geometry-control branch.
LoRA is applied to the attention projections, feed-forward projections, and control-bridge projections of the geometry-control branch with rank 32.
The control bridges are initialized to zero.
The main denoising backbone and video VAE remain frozen.
Training is performed for 10{,}000 iterations on two NVIDIA H200 GPUs using a batch size of 12 and a learning rate of $1\times10^{-5}$.
At inference time, all conditions are inferred directly from fMRI signals.
We use classifier-free guidance with a guidance scale of 5.0 and 20 sampling steps.

\begin{table*}[t]
\centering
\renewcommand\arraystretch{1.0}
\caption{
Subject-wise quantitative results of fMRI2Face on fMRI-Face.
The mean results correspond to the fMRI2Face performance reported in the main paper.
}
\label{tab:sub_results}
\small
\resizebox{1.0\linewidth}{!}{
\begin{tabular}{lccccccccc}
\toprule
Subject
&
PSNR $\uparrow$
&
PixCorr $\uparrow$
&
SSIM $\uparrow$
&
LPIPS $\downarrow$
&
FVD $\downarrow$
&
ID-CSIM $\uparrow$
&
LMD $\downarrow$
&
LMD-A $\downarrow$
&
FWE $\downarrow$
\\
\midrule
Subject 1
& 18.7449
& 0.6822
& 0.8353
& 0.1885
& 78.6234
& 0.3682
& 0.0484
& 0.0833
& 0.0052
\\
Subject 2
& 18.1134
& 0.6388
& 0.8279
& 0.2007
& 82.1460
& 0.3513
& 0.0536
& 0.0866
& 0.0052
\\
Subject 3
& 18.1130
& 0.6337
& 0.8230
& 0.2083
& 87.4255
& 0.3177
& 0.0575
& 0.0891
& 0.0052
\\
\midrule
\textbf{Mean}
& \textbf{18.3238}
& \textbf{0.6516}
& \textbf{0.8288}
& \textbf{0.1991}
& \textbf{82.7316}
& \textbf{0.3458}
& \textbf{0.0532}
& \textbf{0.0863}
& \textbf{0.0052}
\\
\bottomrule
\end{tabular}
}
\end{table*}

\begin{figure*}
  \centering
  \includegraphics[width=0.85\linewidth]{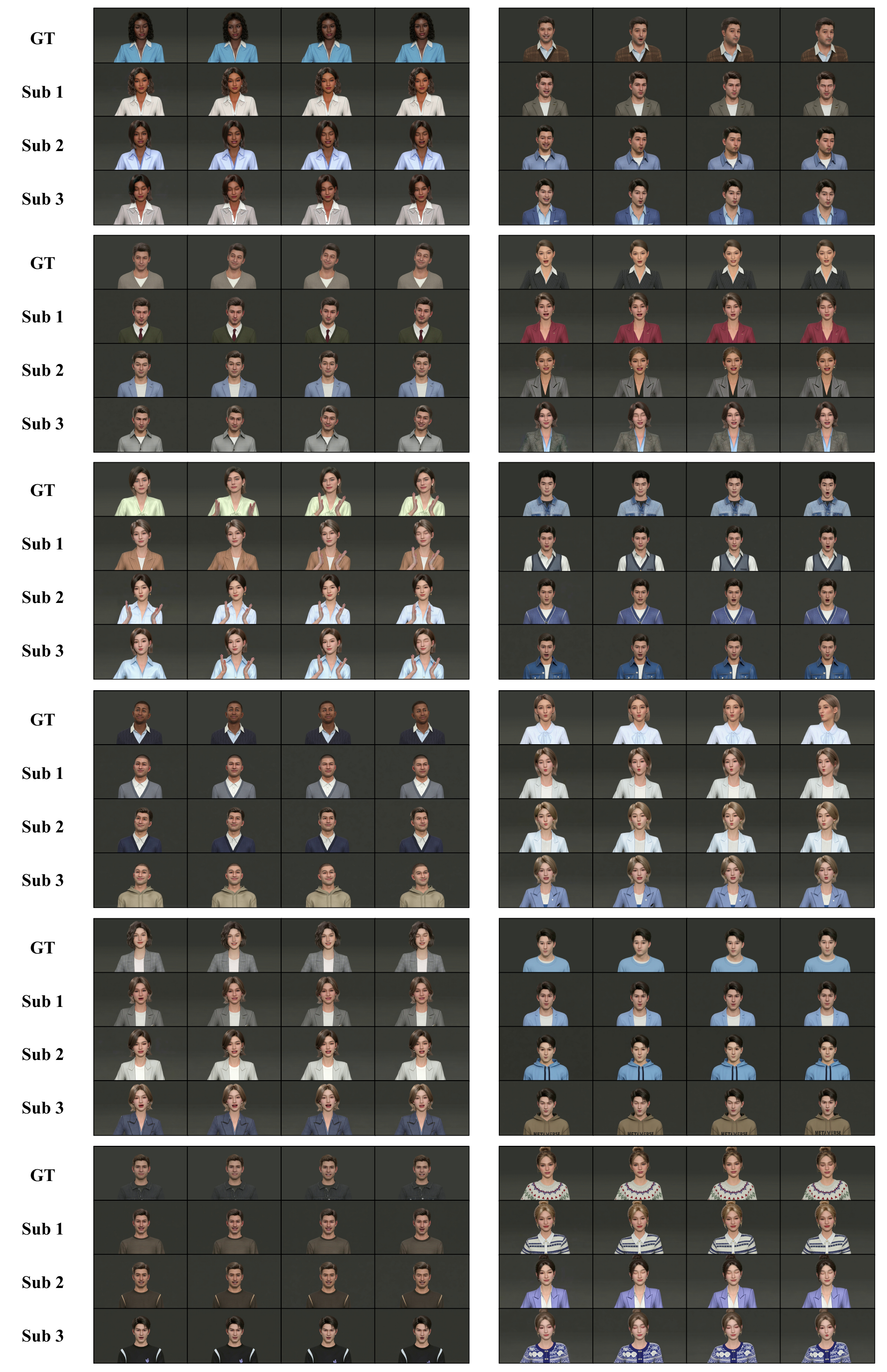}
  \caption{Subject-wise visualization of reconstructed digital human facial videos. 
  }
  \label{fig:sub_vis}
\end{figure*}

\subsection{Additional Metric Details}
\label{sec:supp_metric_details}

We provide additional implementation details of the evaluation metrics.
For landmark-based evaluation, we detect 68 2D facial landmarks from each reconstructed video and compare them with the corresponding ground-truth landmarks.
We report two landmark distances.
LMD is computed in the original image coordinate space, while LMD-A is computed in the aligned face-crop coordinate space following the DECA preprocessing protocol.
Specifically, a similarity transformation is estimated from the facial landmarks to obtain the canonical face crop, and landmarks are normalized in this crop coordinate system.
For both LMD and LMD-A, we compute the mean Euclidean distance over all valid frames and all 68 landmarks:
\begin{equation}
    \mathrm{LMD}
    =
    \frac{1}{KJ}
    \sum_{t=1}^{K}
    \sum_{j=1}^{J}
    \left\|
    \hat{\mathbf{p}}_{t,j}
    -
    \mathbf{p}_{t,j}
    \right\|_{2},
\end{equation}
where $K$ is the number of evaluated frames and $J=68$ is the number of facial landmarks.
The same formula is used for LMD-A, except that the landmark coordinates are measured in the aligned crop coordinate space.

For temporal motion consistency, we report Flow Warping Error (FWE).
For each adjacent frame pair, we estimate the backward optical flow from the ground-truth video and use it to warp the reconstructed previous frame toward the next frame.
FWE is computed as:
\begin{equation}
    \mathrm{FWE}
    =
    \frac{1}{K-1}
    \sum_{t=1}^{K-1}
    \left\|
    \mathcal{W}
    \left(
    \hat{\mathbf{I}}_{t},
    \mathbf{u}^{\mathrm{gt}}_{t+1\rightarrow t}
    \right)
    -
    \hat{\mathbf{I}}_{t+1}
    \right\|_{1},
\end{equation}
where $\mathbf{u}^{\mathrm{gt}}_{t+1\rightarrow t}$ denotes the ground-truth backward optical flow from $\mathbf{I}_{t+1}$ to $\mathbf{I}_{t}$, and $\mathcal{W}(\cdot)$ denotes image warping.
Lower FWE indicates that the reconstructed video follows the temporal motion pattern of the ground-truth video more closely.

\subsection{Subject-specific Results}

We report subject-specific quantitative results in Tab.~\ref{tab:sub_results} and visual examples from the three subjects in Fig.~\ref{fig:sub_vis}.
fMRI2Face produces reconstructions that are visually consistent with the ground-truth videos across all subjects.
Moderate across-subject variation is still observed.
Subject 3 shows slightly weaker quantitative and qualitative performance than Subjects 1 and 2, which may be related to individual differences in neural response patterns, attention stability during scanning, or acquisition noise.



\subsection{Additional Ablation Studies}

\begin{table*}[t]
\centering
\renewcommand{\arraystretch}{1.05}
\setlength{\tabcolsep}{3.0pt}
\caption{
Ablation study on the auxiliary latent representation and 3D control condition.
``Aux. only'' uses only the auxiliary latent without the rendered Morphable 3D Facial Control as the diffusion control condition.
``w/o latent init.'' removes the auxiliary latent initialization stage.
``Frozen aux.'' freezes the pretrained auxiliary latent predictor during diffusion training.
The best results are shown in bold.
}
\label{tab:supp_aux_ablation}
\small
\resizebox{1.0\linewidth}{!}{
\begin{tabular}{lccccccccc}
\toprule
Variant
&
PSNR $\uparrow$
&
PixCorr $\uparrow$
&
SSIM $\uparrow$
&
LPIPS $\downarrow$
&
FVD $\downarrow$
&
ID-CSIM $\uparrow$
&
LMD $\downarrow$
&
LMD-A $\downarrow$
&
FWE $\downarrow$
\\
\midrule
Aux. only
& 17.6044
& 0.5632
& 0.8100
& 0.2099
& 150.1261
& 0.3305
& 0.0585
& 0.0991
& 0.0083
\\
w/o latent init.
& 17.6151
& 0.6168
& 0.8263
& 0.1970
& 94.2275
& 0.3555
& 0.0486
& 0.0840
& 0.0058
\\
Frozen aux.
& 18.6210
& 0.6754
& 0.8332
& 0.1923
& 86.6564
& 0.3618
& 0.0488
& 0.0844
& 0.0053
\\
\midrule
\textbf{Full}
& \textbf{18.7449}
& \textbf{0.6822}
& \textbf{0.8353}
& \textbf{0.1885}
& \textbf{78.6234}
& \textbf{0.3682}
& \textbf{0.0484}
& \textbf{0.0833}
& \textbf{0.0052}
\\
\bottomrule
\end{tabular}
}
\end{table*}

\subsubsection{Ablation on Auxiliary Latent}

We further conduct an auxiliary-latent ablation on subject 1 to analyze its role and interaction with the morphable 3D facial control condition.
\textbf{(1)} As shown in Tab.~\ref{tab:supp_aux_ablation}, using only the auxiliary latent without the rendered morphable 3D facial control leads to clearly worse performance.
This variant obtains a much higher FVD and FWE, indicating that the auxiliary latent alone cannot provide sufficiently explicit geometry-aware motion guidance for facial video reconstruction.
It also shows worse LMD and LMD-A, suggesting that explicit Morphable 3D Facial Control is important for recovering facial structure and pose.
\textbf{(2)} We then evaluate the effect of auxiliary latent initialization.
Without latent initialization, the model already benefits from the rendered 3D facial control, achieving better FVD, LMD, and LMD-A than the auxiliary-only variant.
However, its low-level reconstruction metrics remain lower than the initialized variants, indicating that directly learning the auxiliary latent during diffusion training is less stable.
Pretraining the auxiliary latent predictor before diffusion training improves PSNR, PixCorr, SSIM, LPIPS, and FVD, showing that latent initialization provides a better starting point for recovering visual information beyond the rendered facial region.
\textbf{(3)} Finally, we compare freezing and updating the auxiliary latent predictor during diffusion training.
Freezing the pretrained auxiliary latent predictor already achieves strong performance, but jointly optimizing it with the geometry-control branch yields the best results across all metrics.
This demonstrates that the auxiliary latent predictor benefits from both latent-space initialization and task-specific adaptation within Neural-Controlled Video Diffusion.
Overall, these results confirm that the full design, combining rendered Morphable 3D Facial Control, initialized auxiliary latent completion, and joint diffusion-stage optimization, is necessary for high-fidelity facial video reconstruction.

\begin{table}
\centering
\renewcommand{\arraystretch}{1.0}
\caption{
Comparison of different loss configurations and their corresponding metrics.
Starting from parameter-level supervision, we progressively add landmark reprojection, eye/lip constraints, and temporal motion losses.
The best results are shown in bold.
}
\label{tab:deca_loss}
\small
\resizebox{1.0\linewidth}{!}{
\begin{tabular}{lccccc}
\toprule
Variant
&
PVE $\downarrow$
&
LMD $\downarrow$
&
LMD-A $\downarrow$
&
Eye $\downarrow$
&
Lip $\downarrow$
\\
\midrule
$\mathcal{L}_{\mathrm{param}}$ only
& 0.0195
& 0.0577
& 0.1083
& 0.0168
& 0.0271
\\
+ $\mathcal{L}_{\mathrm{lmk}}$
& 0.0189
& 0.0503
& 0.0875
& 0.0172
& 0.0279
\\
+ $\mathcal{L}_{\mathrm{lip}}$ \& $\mathcal{L}_{\mathrm{eye}}$
& \textbf{0.0188}
& 0.0501
& 0.0873
& 0.0149
& 0.0260
\\
\midrule
+ $\mathcal{L}_{\Delta}$ \textbf{(Full)}
& \textbf{0.0188}
& \textbf{0.0494}
& \textbf{0.0868}
& \textbf{0.0148}
& \textbf{0.0257}
\\
\bottomrule
\end{tabular}
}
\end{table}

\begin{figure*}
  \centering
  \includegraphics[width=0.85\linewidth]{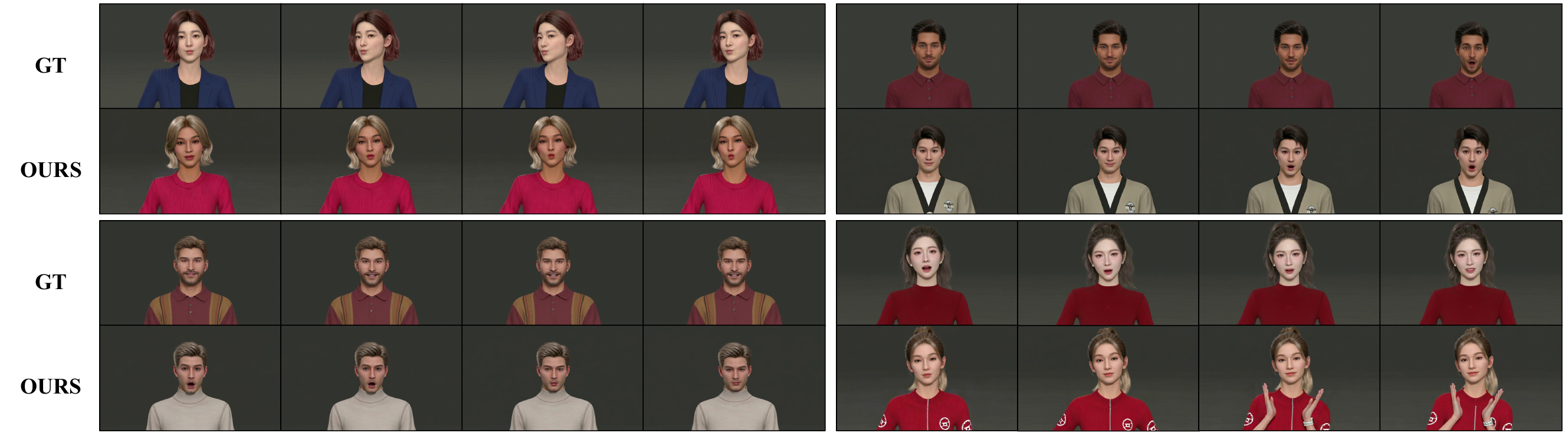}
  \caption{Visualizations for failure cases. 
  }
  \label{fig:fail}
\end{figure*}

\subsubsection{Loss Ablation for 3D Facial Decoding}

We further evaluate the contribution of each loss component in the Morphable 3D Facial Control stream.
Starting from parameter-level supervision only ($\mathcal{L}_{\mathrm{param}}$ only), we progressively add landmark reprojection loss (+ $\mathcal{L}_{\mathrm{lmk}}$), eye/lip constraints (+ $\mathcal{L}_{\mathrm{lip}}$ \& $\mathcal{L}_{\mathrm{eye}}$), and temporal motion loss (+ $\mathcal{L}_{\Delta}$).
As shown in Tab.~\ref{tab:deca_loss}, using only the parameter regression loss provides a reasonable estimate of the DECA parameters but leads to larger reprojection errors.
Adding landmark reprojection supervision substantially improves facial alignment, reducing LMD from 0.0577 to 0.0503 and LMD-A from 0.1083 to 0.0875.
Introducing eye and lip constraints further improves fine-grained facial regions, reducing the eye error and the lip error. 
This suggests that fine-grained regions such as the lips and eyes require dedicated local constraints; without the lip and eye losses, mouth and eyelid movements are poorly represented.
Finally, adding temporal motion losses yields the best overall performance, further reducing landmark errors.
These results show that parameter-level supervision provides the basic 3D facial representation, landmark reprojection improves geometric alignment, fine-grained eye/lip losses refine local facial details, and temporal motion losses enhance dynamic facial consistency.

\subsection{Failure Cases}

We present several representative failure cases in Fig.~\ref{fig:fail}. 
In the first two rows, the reconstructed videos exhibit inconsistencies in global appearance. For example, the female digital human with red hair is reconstructed as a blonde individual wearing red clothing, reflecting ambiguity in associating color cues with the correct appearance attributes. In the third and fourth rows, the main errors arise from facial motion reconstruction, where the predicted expression or head movement deviates from the ground truth.

\section{Limitations and Future Work}

Although fMRI2Face achieves high-fidelity facial video reconstruction, several limitations remain.
First, reconstructing fine-grained transient facial motions remains challenging.
Although the proposed geometry-guided design improves facial structure, pose, expression, and motion consistency, rapid micro-motions such as eye blinks or very subtle expression transitions are difficult to recover reliably from fMRI.
This limitation is partly due to the relatively low temporal resolution of fMRI and the delayed hemodynamic response, which undersample fast neural dynamics.
Future studies may combine fMRI with brain recording techniques of higher temporal precision, or introduce stronger temporal priors for micro-expression modeling.

Second, fMRI-Face uses high-fidelity digital human stimuli to balance realism and experimental controllability.
This design enables large-scale controlled data collection and systematic analysis of identity, expression, and head motion.
However, digital human stimuli still differ from fully natural real-world facial videos in terms of environmental complexity, spontaneous behavior, and social context.
Future work could extend fMRI-Face with more diverse identities, richer motions, more naturalistic interactions, and additional participants.
Such extensions would help further evaluate the generality of fMRI-based facial video reconstruction and deepen our understanding of dynamic face perception in the human brain.

\section{Broader Impact}

The fMRI-Face dataset and fMRI2Face framework may benefit several research areas.
First, they provide a new benchmark for dynamic face decoding from brain activity. 
Compared with prior face-decoding datasets that mainly focus on static or less controllable stimuli, fMRI-Face incorporates controlled, continuous facial motion, enabling the study of how the brain processes dynamic social signals over time.
This opens new possibilities for modeling and reconstructing perceived video content from fMRI activity.
Second, fMRI-Face can support neuroscience research on face perception and social cognition.
Because the stimuli are generated under controlled rendering conditions, researchers can examine how specific visual factors, such as identity-related appearance, expression dynamics, and head motion, are reflected in neural responses.
The dataset may therefore help study the functional organization of visual and face-selective regions, as well as the neural encoding of dynamic social cues.
Third, the proposed framework connects neural decoding, controllable digital humans, and generative video modeling.
By integrating Brain-derived Appearance Context, Morphable 3D Facial Control, and Neural-Controlled Video Diffusion, fMRI2Face provides a structured way to study how brain activity can be mapped to both global appearance and geometry-aware facial motion.
We hope this will encourage future work on more interpretable and controllable neural decoding systems.

At the same time, brain decoding technologies require careful and responsible use.
Future applications of neural decoding should follow strict ethical guidelines, including informed consent, privacy protection, secure data handling, and clear communication of technical limitations.

\end{document}